\documentclass[journal]{IEEEtran}

\usepackage{cite}
\usepackage[pdftex]{graphicx}
\usepackage{amsmath}
\makeatletter
\def\maketag@@@#1{\hbox{\m@th\normalfont\normalsize#1}}
\makeatother
\usepackage{algorithmic}
\usepackage{array}
\usepackage{url}
\usepackage{placeins}
\usepackage{mathtools}
\usepackage{nccmath}
\usepackage{amssymb}
\usepackage{hyperref}
\usepackage{bbm}
\usepackage{bigints}
\usepackage{xcolor} 
\usepackage{marginnote}
\usepackage{placeins}

\DeclareMathOperator*{\argmin}{argmin}

\newcommand{\indep}{\rotatebox[origin=c]{90}{$\models$}}

\newcommand{\rebmod}[1]{%
	#1
}
\newcommand{\rebnew}[1]{%
	#1
}

\begin{document}

%\title{On the Creation of a semi-synthetic Dataset for Image Segmentation based on Real Images: \\ Robotic Instrument Labelling as a Pretext Task}
%\title{Creating a Semi-synthetic Dataset for Segmentation: Robotic Instrument Labelling as a Pretext Task}
\title{
%Image Compositing for Semantic Segmentation 
Image Compositing for Segmentation of Surgical Tools without Manual Annotations
%Robotic Instrument Segmentation
%Image Compositing for Learning Robotic Instrument Segmentation from Synthetic Data
}

\author{Luis~C.~Garcia-Peraza-Herrera, %~\IEEEmembership{Student Member,~IEEE,}
        Lucas~Fidon, %~\IEEEmembership{Student Member,~IEEE,}
        Claudia~D'Ettorre, %~\IEEEmembership{Student Member,~IEEE,}
        Danail~Stoyanov, %~\IEEEmembership{Member,~IEEE,}
        Tom~Vercauteren, %~\IEEEmembership{Member,~IEEE,}
        and~S\'ebastien~Ourselin %~\IEEEmembership{Member,~IEEE}% <-this % stops a space
\thanks{Paper submitted for review on \today. This work was supported by Wellcome [203148/Z/16/Z; 203145Z/16/Z; WT101957], EPSRC [NS/A000049/1; NS/A000050/1; NS/A000027/1; EP/L016478/1]. This project has received funding from the European Union's Horizon 2020 research and innovation program under the Marie Sk{\l}odowska-Curie grant agreement TRABIT No 765148. Tom Vercauteren is supported by a Medtronic / RAEng Research Chair [RCSRF1819\textbackslash7\textbackslash34]. The authors would like to thank NVIDIA for the donated GeForce GTX TITAN X GPU.}
\thanks{L. C. Garcia-Peraza-Herrera, C. D'Ettorre and D. Stoyanov are with University College London, United Kingdom (e-mail: luiscarlos.gph@gmail.com). L. C. Garcia-Peraza-Herrera, L. Fidon, T. Vercauteren and S. Ourselin are with King's College London, United Kingdom.}
%\thanks{Manuscript received Month DD, 2019; revised Month DD, 2019.}
}

% note the % following the last \IEEEmembership and also \thanks - 
% these prevent an unwanted space from occurring between the last author name
% and the end of the author line. i.e., if you had this:
% 
% \author{....lastname \thanks{...} \thanks{...} }
%                     ^------------^------------^----Do not want these spaces!
%
% a space would be appended to the last name and could cause every name on that
% line to be shifted left slightly. This is one of those "LaTeX things". For
% instance, "\textbf{A} \textbf{B}" will typeset as "A B" not "AB". To get
% "AB" then you have to do: "\textbf{A}\textbf{B}"
% \thanks is no different in this regard, so shield the last } of each \thanks
% that ends a line with a % and do not let a space in before the next \thanks.
% Spaces after \IEEEmembership other than the last one are OK (and needed) as
% you are supposed to have spaces between the names. For what it is worth,
% this is a minor point as most people would not even notice if the said evil
% space somehow managed to creep in.

% The paper headers
%\markboth{IEEE Transactions on Medical Imaging,~Vol.~XX, No.~X, April~2019}%
%{Shell \MakeLowercase{\textit{et al.}}: Bare Demo of IEEEtran.cls for IEEE Journals}
\markboth{Garcia-Peraza-Herrera \MakeLowercase{\textit{et al.}}}%
{Garcia-Peraza-Herrera \MakeLowercase{\textit{et al.}}}
% The only time the second header will appear is for the odd numbered pages
% after the title page when using the twoside option.
% 
% *** Note that you probably will NOT want to include the author's ***
% *** name in the headers of peer review papers.                   ***
% You can use \ifCLASSOPTIONpeerreview for conditional compilation here if
% you desire.

% If you want to put a publisher's ID mark on the page you can do it like
% this:
%\IEEEpubid{0000--0000/00\$00.00~\copyright~2015 IEEE}
% Remember, if you use this you must call \IEEEpubidadjcol in the second
% column for its text to clear the IEEEpubid mark.

% use for special paper notices
%\IEEEspecialpapernotice{(Invited Paper)}

% make the title area
\maketitle

% As a general rule, do not put math, special symbols or citations
% in the abstract or keywords.
\begin{abstract}
Producing manual, pixel-accurate, image segmentation labels is tedious and time-consuming.
This is often a rate-limiting factor when large amounts of labeled images are required, such as for training deep convolutional networks for instrument-background segmentation in surgical scenes. 
No large datasets comparable to industry standards in the computer vision community are available for this task.
To circumvent this problem, we propose to automate the creation of a realistic training dataset by exploiting techniques stemming from special effects and harnessing them to target training performance rather than visual appeal.
Foreground data is captured by placing sample surgical instruments over a chroma key (a.k.a. green screen) in a controlled environment, thereby making extraction of the relevant image segment straightforward. 
Multiple lighting conditions and viewpoints can be captured and introduced in the simulation by moving the instruments and camera and modulating the light source.
Background data is captured by collecting videos that do not contain instruments. In the absence of pre-existing instrument-free background videos, minimal labeling effort is required, just to select frames that do not contain surgical instruments from videos of surgical interventions freely available online.
We compare different methods to blend instruments over tissue and propose a novel data augmentation approach that takes advantage of the plurality of options.
We show that by training a vanilla U-Net on semi-synthetic data only and applying a simple post-processing, we are able to match the results of the same network trained on a publicly available manually labeled real dataset.
\end{abstract}

\begin{IEEEkeywords}
Image Compositing, Chroma Key, Tool Segmentation.
\end{IEEEkeywords}
\IEEEpeerreviewmaketitle

\section{Introduction}
\IEEEPARstart{S}{urgical} instrument labeling is a generic problem in Computer-Assisted Interventions (CAI) \cite{Rieke2016a, Ye2016, Garcia-Peraza-Herrera2017c, Pakhomov2017a, Shvets2018, Nwoye2018}. Locating an instrument's contour within a surgical scene
has a wealth of potential applications.
%could be potentially used in a substantial variety of tasks.
It is already, or in some cases is bound to become, an essential building block of many %computer-assisted surgical 
key clinical applications in Surgical Data Science \cite{Maier-Hein2017b}.
To name a few: placing informative overlays on the screen or performing augmented reality without occluding instruments, subtracting surgical tools from the scene when building a tissue panorama,
%as the endoscope moves, 
surgical workflow analysis, skills analysis and error detection, automatic endoscope camera calibration, visual servoing, surgical task automation, feature matching for 3D reconstruction, and in general any approach that takes advantage of real-time segmentation as a way of tracking a target across frames.
\begin{figure}[!t]
	\centering
	\includegraphics[width=\columnwidth]{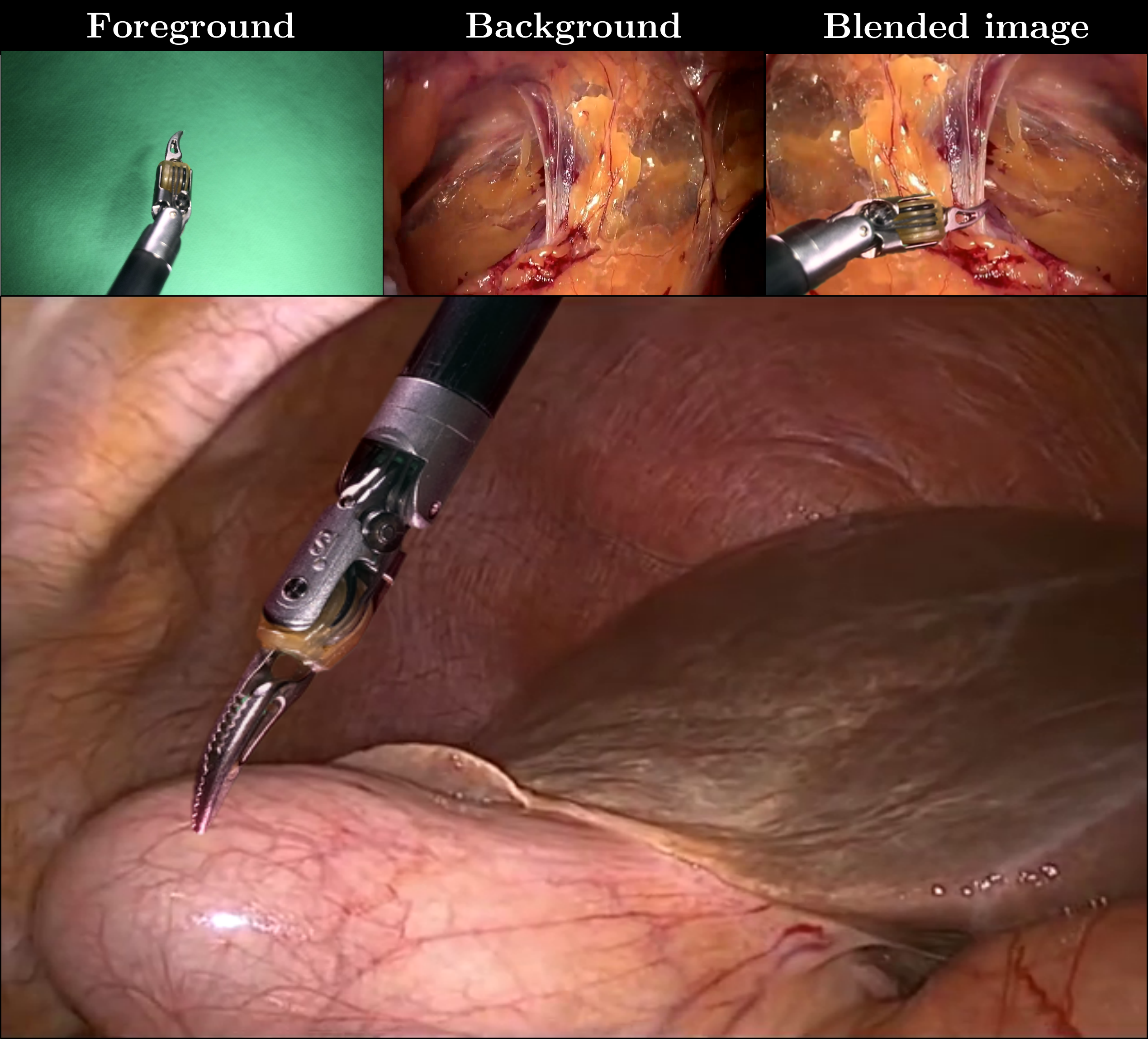}
	\caption{
		Blending process (top row). Blended image sample (bottom picture). Inspired by chroma key compositing in the movie industry, we show that segmentation models can be trained exclusively on a semi-synthetic dataset based on superimposition and data augmentation %(scaling, rotation, and flip, 
		(as shown in the blended image above), achieving a comparable accuracy to those trained on manually annotated images. We demonstrate that the difficulty to synthesize a realistic border represents less of a problem for learning purposes when our stochastic mix of known blending methods (called \textit{mix-blend}) is employed to superimpose objects, allowing for state-of-the-art segmentation performance.
	}
	\label{fig:thief_of_bagdad}
\end{figure}

Early approaches for tool recognition either used positioning sensors (robotic surgery) \cite{Kwartowitz2009} or embedded additional sensors \cite{Haase2013} within the instruments. However, attempts to do so have shown many drawbacks besides a limited accuracy \cite{Reiter2012, Rocha2019}. In addition to the difficulties created by the added complexity due to the workflow alteration, additional sensors or tool modifications have to be able to overcome the harsh conditions of the instrument sterilization process \cite{Pakhomov2020}.

Recent machine learning advances have shown extraordinary progress in visual recognition. Yet, in the medical context in general, and for our task in particular, progress is typically restrained by the scarcity of available fine-grained annotations required for training purposes, and by the limited practicality and cost of creating such annotations at scale. In this paper, we explore the feasibility of using a chroma key (see fig. \ref{fig:thief_of_bagdad} and supplementary material fig. $2$ and $3$) to automatically generate large quantities of semi-synthetic yet realistic ``ground truth" images and labels. As opposed to synthetic data, our generated images come from compositing real images. Hence, we refer to them as semi-synthetic. We use them to train a deep neural network for surgical tool segmentation.
Given the proposed setup in fig. \ref{fig:thief_of_bagdad}, the key methodological challenge is the blending of instruments recorded over the chroma key onto tissue frames in such a way that the segmentation learned based on this semi-synthetic images generalizes to real clinical data. 
Stemming from this methodological challenge, it is a research question whether it is necessary to blend realistic images for learning the segmentation, or in contrast, it can be learned without the need for deploying sophisticated domain-specific blending mechanisms.
%- Is there an ideal blending?
%- Does the blending have to look realistic?
%- Is one way of blending enough?
%- Can we learn the segmentation employing only a set of simple but slightly inacurate set of blending modes 

The paper is structured as follows. First, we comment on recent related articles that target different forms of image compositing to learn different tasks. We continue to explain our formalization of the learning problem and how 
%we model endoscopic images and 
this leads to the generation of the semi-synthetic training instances used to learn the segmentation. Then, we introduce a training strategy that adapts to our modelling of the images and way of blending. The material's section contains a detailed description of the data used for the experiments and how we record it. In the implementation section, we develop on data augmentation, specific blending modes employed, and network training protocol. Finally, we explain the different experiments carried out to evaluate the performance of our methods, and discuss the results obtained.

\textit{Contributions}. We propose a novel technique and theoretical framework to synthesize ground truth images and labels for image segmentation problems.
We focus on instrument segmentation in endoscopic scenes.
Our method relies on two sources of data: Sample instruments recorded over a chroma key; and images that only contain background tissue.
%
%We explore how to recreate surgical scene conditions over a chroma key such that the semi-synthetic ground truth allows our segmentation network to generalize to real scenes  
%
In order to merge these two pieces of information we rely on existing blending methods and propose \textit{mix-blend}. This novel blending approach relies on the probabilistic combination of a set of simple blending functions that act as a basis for blending. Furthermore, we introduce a Monte Carlo method to generate the blended training samples on the fly (during training), which fits well with Stochastic Gradient Descent (SGD) optimization solvers.
This approach allows us to learn the segmentation without the need for 
advanced
%exotic
domain adaptation techniques.
We also make public all our newly created datasets including: our semi-synthetic tool segmentation dataset containing 100K labeled images, our chroma key foreground dataset (14K labeled images), our background tissue dataset (6K images from 50 surgical procedures), %our unlabeled domain adaptation real clinical dataset (160K images from 50 surgical procedures), 
and our real clinical testing set (514 manually labeled images from 20 surgical procedures).
\begin{figure*}[htb!]
    \centering
	\includegraphics[width=\textwidth]{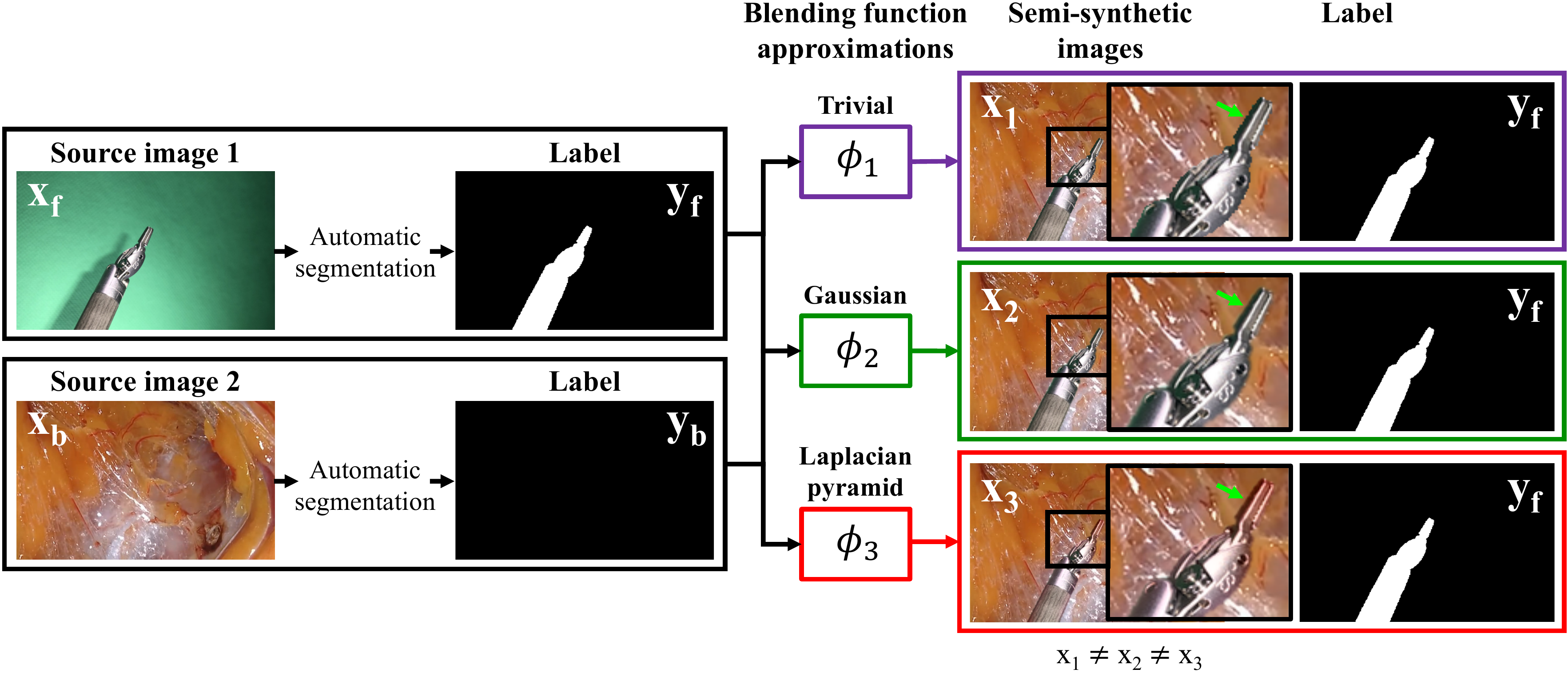}
	\caption{Overview of the semi-synthetic generation method. 
	We illustrate both the concept of \textit{source image} and the process to generate semi-synthetic images. For an image to be considered as \textit{source image}, it must fulfill two conditions. First, its labeling must be available or easier than a manual annotation of the tool in a real endoscopic image. Second, it must be a close approximation of a real clinical image so that when complementary \textit{source images} are blended, the result is close to a real endoscopic image. The automatic segmentation of \textit{source images} is hue-based for $\boldsymbol{x_f}$ (detailed in section \ref{subsection_foreground_dataset_segmentation}) and trivial for $\boldsymbol{x_b}$. We assume that the correct way to blend complementary source images to form a real clinical image is unknown. In this figure, three possible ways to do it are illustrated. A green arrow is shown to highlight a part of the semi-synthetically generated image, the border of the tool, and the tooltip itself, whose appearance differs depending on the choice of blending function (best viewed in the electronic version of the manuscript).}
	\label{fig:modelling_of_x}
\end{figure*}
\section{Related Work}
A number of classical surgical tool segmentation methods have been proposed with promising results, see e.g. \cite{Rieke2016a,Ye2016}.
%In \cite{Rieke2016a}, the authors propose a Random Forest template matching of intensity-based features to detect a bounding box around the instrument. Similarly, \cite{Ye2016} divides instruments in $14$ parts and tracks them using template matching. Manual initialization of the initial instrument position is required.
Recently, data-driven approaches based on Convolutional Neural Networks \cite{Garcia-Peraza-Herrera2017c, Pakhomov2017a, Shvets2018} have shown to be the leading technique to estimate an instrument-background segmentation of surgical scenes \cite{Allan2019}.
Despite the availability of such powerful methods,
%changing lighting conditions, and varying viewpoint and resolution (of objects and scenes)
the tool segmentation task
still represents a challenge due to a lack of large-scale fine-grained annotated datasets.
As pointed out in the conclusions of the 2017 Robotic Instrument Segmentation Challenge \cite{Allan2019}, current instrument segmentation datasets, such as the one used in the challenge (only around $3$K annotated images), are severely limited by the amount of data. In contrast, general-purpose computer vision datasets possess hundreds of thousands or millions of images (e.g. COCO, 330K manually segmented images). 
In order to circumvent the need for similar quantities of manual annotations, recent works have explored ways of reducing human effort. This has been of interest in both the computer vision industry \cite{Zhou2015, Dwibedi2017, Georgakis2017, Dvornik2018, Tripathi2019} and the CAI community \cite{Heimann2014, Maier-Hein2016, Vardazaryan2018, Nwoye2018, Ross2018}.
There are two fundamental approaches to alleviate the manual effort. Either reducing the number of annotations needed or allowing for faster manual labeling. In the following paragraphs, we give an overview of both.

In \cite{Maier-Hein2016}, Maier-Hein et al. explored crowd-sourcing as a means of correcting weak labels generated from a small amount of annotated images (\textit{MICCAI Endoscopic Vision Challenge 2015}). In their proposal, workers were provided with endoscopic images and corresponding estimated segmentations, which they had to correct with an interactive tool. The estimated segmentations were generated by an atlas forest (AF) method \cite{Zikic2013a}. The authors trained an uncertainty estimator based on the predictions of the forest to regress confidence maps, providing the workers with only those areas of low confidence for correction. 
Although the results are encouraging, this method still requires an interactive segmentation setup and manual pixel-wise annotations. 
% other authors have explored how to use classification labels to train networks able to produce heatmaps whose blobs give an approximate localization of the different objects in the image. 
%from the target domain during training. 
%
In another attempt to reduce the labeling effort, Nwoye et al. \cite{Nwoye2018} have recently suggested to use per-frame instrument presence classification labels as weak supervision for segmentation.
%In their work, they train a convolutional network using these annotations and obtain heatmaps whose blobs provide an approximated localization of the tools in the image.
Although they have shown promising performance for localization, 
%which is useful for tasks such as surgical phase segmentation, 
it is still very challenging for these methods to provide a pixel-wise accurate segmentation of the instruments, which is crucial for applications such as visual servoing for surgical task automation. Besides, a labeled dataset for tool presence is mandatory.
Another approach to alleviating the manual labeling effort is to employ unlabeled data to learn the endoscopic image domain representation. Ross et al. \cite{Ross2018} proposed to learn re-colorization as a pre-training task. They convert endoscopic images to CIELab and train a convolutional network in a fake/real adversarial scheme to regress the colour channels \textit{a} and \textit{b}, giving L as an input. This auxiliary task allows them to reduce the amount of manually annotated data considerably. % required to achieve a comparable accuracy.
While showing promising results, pixel-wise annotations are still required.
Alternatively to pre-training on a different auxiliary task, Yu et al. \cite{Yu2018} proposed to reduce the amount of labeling by exploiting temporal consistency present in surgical videos.
%pre-training on the same task. The novelty versus previous approaches being the use of information for pre-training that is not available at inference time. %
In their work, a teacher network uses past and future video frames to produce a classification estimation for the \textit{current} video frame. The teacher network is then employed to generate a larger dataset of weak labels based on unlabeled data, from which the student network learns, using the same information as input that is employed at testing time.
%
%Another approach that relies on weak annotations have been introduced by Fuentes-Hurtado et al. \cite{Fuentes-Hurtado2019}. In their proposal, they suggest to manually label images for segmentation with straight lines over the objects of the different classes as opposed to the usual time-consuming full annotations. While encouraging, as labeling time is shortened, and results show that performance is close to training with fully annotated labels, per-frame manual labeling is still required.

%As an alternative to reducing the number of labels or time required for dataset curation, dataset synthesis has emerged as an inexpensive approach to generate annotated images automatically.

Recently, generative methods \cite{Pakhomov2020, Pfeiffer2019} have shown potential to address the data scarcity problem in endoscopic vision. Alternatively to just reducing the number of labels or time required for dataset curation, dataset synthesis has emerged as an inexpensive approach to generate annotated images automatically.
The work of Heimann et al. \cite{Heimann2014} proposed to synthetically embed an ultrasound (US) transducer into fluoroscopy images to generate a training set. The aim was to detect the in-plane probe's position, orientation, and scale. To do so, the authors performed a computer tomography (CT) of a US probe and embedded it into real CT patient volumes, automatically generating radiographs and ground truth by a forward projection of the CT that contains the synthetically embedded US probe. A similar approach has been recently presented in \cite{Unberath2019}, where authors show different aspects of the simulation that are key to bridge the gap between synthetic and real data. 

In a non-medical context, there is an established research direction that aims to overcome the lack of annotated data using the so-called virtual setups or synthetically rendered scenes \cite{Su2015a}. This approach is extremely challenging \cite{Peng2015, Chen2016a}. As suggested by \cite{Movshovitz-Attias2016}, many aspects have to be taken into account to synthesize high-quality scenes, often requiring advanced domain adaptation techniques to bridge the gap between synthetic and real data.
To address the generalization difficulty addressed by virtually rendered setups, a new research stream focuses on compositing as opposed to rendering \cite{Karsch2011, Dwibedi2017, Georgakis2017, Dvornik2018, Tripathi2019}. In this line of research, training images are composed by a combination of visual elements coming from different sources.
Augmentation techniques related to compositing, such as \emph{mixup} where pairs of images are alpha-blended \cite{Zhang2017}, have also been developed to improve generalization.
%to create the illusion that the composed image is a real scene. 
%
In \cite{Karsch2011}, authors proposed a method to embed synthetic objects of known geometry into real pictures.
Although they manage to generate photo-realistic semi-synthetic images, their approach relies on a fine-grained model of the 3D scene geometry and the lighting conditions, which is not available in endoscopic videos.
%To do so, they build a physical model of the scene. Manual annotations of the background image geometry (bounding, extruded and occluded) and lighting are required prior to the composition. The realism of this method is superlative, but manual annotations that are not available in endoscopic scenes are still needed for every frame. 

Orthogonal to those techniques that aim to maximize the realism of training data, domain randomization \cite{Tobin2017} has emerged as a powerful technique to bridge the gap between simulation and target domain by doing exactly the opposite. The aim is to alter the synthetic data in a stochastic manner such that the deep networks concentrate on the essential features to solve the task. Reality is modelled as yet another variation of the source domain.
%Moving in this direction, Peng et al. \cite{Peng2018} proposed to learn a deep reinforcement learning policy to control the movement of a robot to perform a pushing task. In order to generalize to a real environment, they randomly sampled the dynamics of the system during the training in the simulator. This allowed them to learn policies that are robust to different environments, including real scenarios.
In medical imaging, domain randomization has recently also shown promising results \cite{Toth2019}. Toth et al. have successfully used this technique for 3D/2D cardiac model-to-X-ray registration, showing that unrealistic perturbations of the training data are useful to train a model for the task just based on synthetic training data.
%The task of choice being how to push a puck over a surface to bring it from its current position to a designated location.

In computer vision, image compositing is one of the possible approaches to perform domain randomization. This has been recently shown by several authors \cite{Dwibedi2017, Tremblay2018}. In \cite{Dwibedi2017}, Dwibedi et al. proposed to use automatic compositing to build a dataset for training a deep learning object detector. %The authors used RGB-D images of objects from the BigBIRD dataset \cite{Singh2014}. Every object in the dataset is placed inside an open box with white background where \mbox{RGB-D} pictures are taken from different points of view. As each picture contains a single object, the authors used the depth map (i.e. depth channel of RGB-D) as segmentation ground truth to train a fully convolutional network. This network is then used by the authors to segment their own pictures, which are taken in a similar setup to the one of BigBIRD. 
Even though image compositing approaches seem promising, they suffer from a recurrent issue, the unintended embedding of artificial features derived from the blending process into the synthetic images.
%and impedes the  to real images~\cite{Dwibedi2017}. 
This creates a bias in the dataset. That is, if all the objects that we are trying to detect are blended using the same method, let us say, a crude cut-and-paste, they are all subject to have a particular boundary derived from the cut-and-paste process. 
%A convolutional network can extract these blending-related features %that are present in all our synthetically superimposed objects as a discriminative feature 
%to detect the objects. 
As objects in real images do not present these features, generalization is highly affected~\cite{Dwibedi2017}.
To alleviate this, Dwibedi et al. \cite{Dwibedi2017} propose to synthesize every training image several times, using the same objects and background, but a different method to superimpose the objects. This way of modelling the blending, and the learning problem it leads to, represent a particular case of the formulation we introduce in section \ref{section_methods} for the problem. We compare this method to ours and propose an alternative approach to model the combination of blending algorithms.
%We formalize their approach, compare it to our results, and propose an alternative way to model the combination of blending algorithms.
%In our paper, we formalize this approach, giving a theoretical insight on this way of modelling the blending and formulating the learning process it leads to. 
%
Although the approach in \cite{Dwibedi2017} shows an improvement of accuracy over a trivial cut-and-paste, it is still limited by the $\mathbb{N}_{>1}$ deterministic blending methods chosen, whose features can also be learned by a deep network. Tripathi et al. \cite{Tripathi2019} propose an alternative approach to prevent the network from exploiting blending artifacts to detect foreground objects. They blend all the objects with a single method, the standard alpha-blending, but introduce artifacts in the background. These artifacts (called flying distractors) consist of parts of other backgrounds, cut with the shape of a foreground object, and blended into the scene. This approach also fits well within our theoretical framework in section \ref{section_methods}, as we are not limited to use a foreground-background image pair to create our semi-synthetic images, but can also use two backgrounds with one of them having the segmentation annotation of another foreground image, leading to the solution proposed by \cite{Tripathi2019}. Hence, flying distractors are also included in our semi-synthetic dataset, blending them with our Monte Carlo method to generate training samples.
\section{Methods}
\label{section_methods}

%%%%%%%%%%%%%%%%%%%%%%%%%%%%%%%%%%%%
\subsection{Semi-synthetic learning}
%%%%%%%%%%%%%%%%%%%%%%%%%%%%%%%%%%%%
%
%\reversemarginpar
%\marginnote{$R_1C_1$}
%\textcolor{orange}{
%  The problem  of XYZ has been studied in.
%}
%
As it is customary in data-driven segmentation, we aim to solve for a mapping $f$ such that $f_{\boldsymbol{\theta}}(\boldsymbol{x}) \approx \boldsymbol{y}$, where $\boldsymbol{x}$ is an input image, $\boldsymbol{y}$ the segmentation mask corresponding to $\boldsymbol{x}$, and $\boldsymbol{\theta}$ a vector of parameters. %As usual in machine learning, 
The parameters $\boldsymbol{\theta}$ are sought as an approximate solution of the following Expected Risk Minimization problem:
\begin{equation}
    \boldsymbol{\hat{\theta}} = \underset{\boldsymbol{\theta}}{\argmin}
    %\mathop{{}\mathbb{E}}{}_{(X, Y) \sim P_{X, Y}}
    \mathop{{}\mathbb{E}}{}_{P_{X, Y}}
    [\ell(f_{\boldsymbol\theta}(X), Y)]
    \label{eq:expected_risk_minimization}
\end{equation}
where $f_\mathbf{\theta}$ is a parametric function (the segmentation network in our implementation),
$\ell(\cdot, \cdot)$ represents our real-valued loss function, and $X, Y$ are modelled as two random variables of unknown joint probability distribution $P_{X, Y}$.
For the sake of simplicity, as we focus on how our modelling of $X$ changes the optimization problem, regularization terms are omitted across this section.

In practice, limited by a training set, we typically approximate the true joint probability distribution $P_{X, Y}$ of our training set by the following empirical probability measure $\hat{P}_{X, Y}$:
\begin{equation}
    \hat{P}_{X, Y}(\boldsymbol{x}, \boldsymbol{y}) = \frac{1}{N}\sum_{n} \delta_{\boldsymbol{x}_{n}}(\boldsymbol{x}) \delta_{\boldsymbol{y}_{n}}(\boldsymbol{y})
    \label{eq:deltas}
\end{equation}
where $n \in \{1, ..., N\}$, $N$ represents the number of training samples, and $\delta_{\boldsymbol{x}_n}$ and $\delta_{\boldsymbol{y}_n}$ stand for Dirac measures centered at $\boldsymbol{x}_n$ and $\boldsymbol{y}_n$ respectively. 
Based on \eqref{eq:deltas}, the learning problem~\eqref{eq:expected_risk_minimization} becomes:
\begin{equation}
\begin{split}
    \boldsymbol{\hat{\theta}} &= \underset{\boldsymbol{\theta}}{\argmin}
    %\mathop{{}\mathbb{E}}{}_{(X, Y) \sim \hat{P}_{X, Y}}
    \mathop{{}\mathbb{E}}{}_{\hat{P}_{X, Y}}
    \Big[\ell\big(f_{\boldsymbol\theta}(X), Y\big)\Big] \\
    %\boldsymbol{\hat{\theta}}
    &= \underset{\boldsymbol{\theta}}{\argmin} 
    \frac{1}{N}
    \sum_{n} 
    \ell\big(f_{\boldsymbol\theta}(\boldsymbol{x}_n), \boldsymbol{y}_n\big)
\end{split}
\end{equation}
%
%where $(\boldsymbol{x}_i, \boldsymbol{y}_i)~\epsilon~\mathcal{X}\times\mathcal{Y}$. 
However, generating pairs $(\boldsymbol{x}, \boldsymbol{y})$ by recording real clinical images and manually labeling is exceedingly time-consuming, leads to inaccurate labels and amounts of data that are far from computer vision industry standards. 
Nonetheless, we observe that in our problem of instrument-background segmentation, the foreground of an image could be overlaid onto the background of another, and still form a plausible image. This gives us the intuition that an image could be segmented into several components (including but not limited to foreground and background) that could be blended to form new images.
%
%That is, $\boldsymbol{x}$ could be 
As such, we consider an image $\boldsymbol{x}$ as a
realization of a random variable $X$ modelled as $X = \phi^{*}(H_1, ... , H_k)$ where $\{H_k\}_{k=1}^{K}$ are $K$ random variables
capturing the different components of information (e.g. background, foreground instrument).
Observations (called \textit{source images} throughout the text) of these random variables represent the \textit{sources} of any given $\boldsymbol{x}$, and $\phi^{*}$ is an ideal blending function that renders the final image with all its components.
This model is particularly advantageous if:
\begin{enumerate}
    \item Labeling samples of $\{H_k\}_{k=1}^{K}$ is easier than labeling samples of $X$.
    \item We are able to blend segmented complementary \textit{source images} and their labels to form new valid training pairs $(\boldsymbol{x}, \boldsymbol{y})$.
\end{enumerate}
%
%A convenient way to segment $\boldsymbol{x}$ is by assuming 
The idea of seeing an endoscopic image as made of $K$ \textit{source images} that are easier to segment than real clinical images themselves is illustrated in fig. \ref{fig:modelling_of_x}. In our case, we assume that we are able to segment any given $\boldsymbol{x}$ into $K$ \textit{source images}, where $K$ is the number of classes. In the case of binary tool segmentation, we focus on $K=2$, foreground (surgical instruments) and background (tissue). Furthermore, we hypothesize that any combination of complementary \textit{source images} is equally likely to form a plausible endoscopic image. This proposal corresponds to modelling $X$ as:
\begin{equation}
    \begin{split}
        &X = \phi^{*}(X_F, X_B) \\
        &X_F \indep X_B
        \label{eq:model_x}
    \end{split}
\end{equation}
where $X_F := H_1$ and $X_B := H_2$ are assumed to be independent random variables whose observations are foregrounds $\boldsymbol{x_f}$ and backgrounds $\boldsymbol{x_b}$. 
\begin{figure*}[!t]
    \centering
	\includegraphics[width=\textwidth]{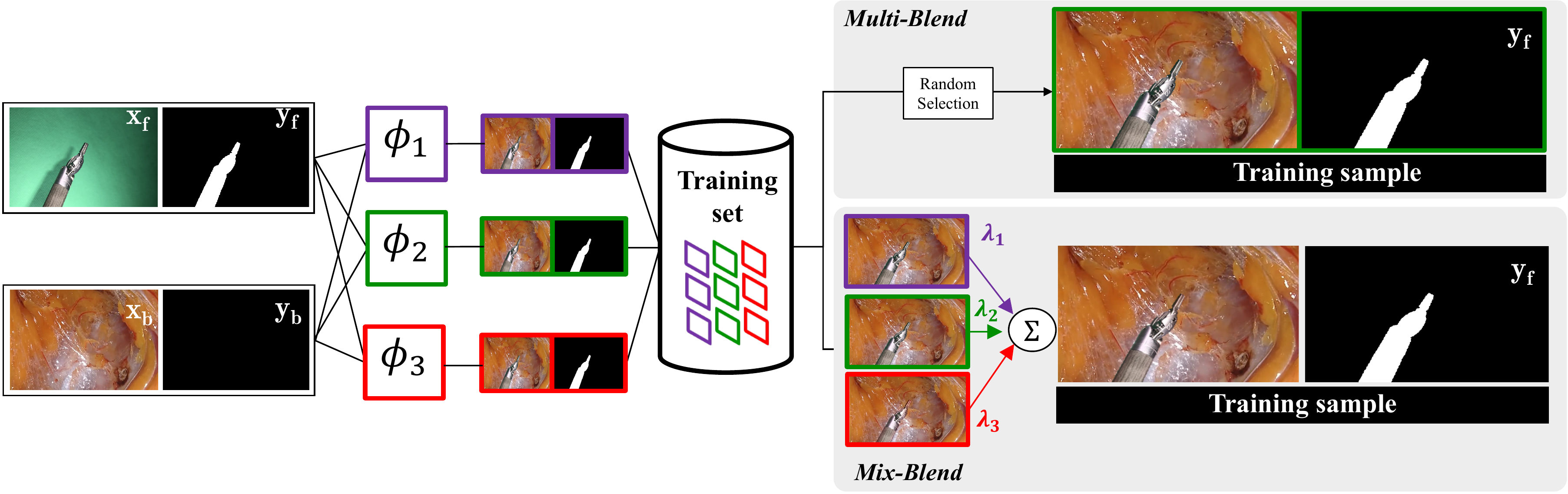}
	\caption{
	\textit{Multi-blend} training iteration: in every training iteration a pair foreground-background is chosen and blended with one of the $M$ blending functions we decide to employ. To comply with \eqref{eq:all_images_learning}, all $M$ blending functions have to be chosen at least once during the training for each combination of $(\boldsymbol{x_f}, \boldsymbol{x_b})$ present in the dataset. After a training instance (image plus label) has been generated, it is passed to $f_{\boldsymbol{\theta}}$ (see eq. \ref{eq:expected_risk_minimization}).
	\textit{Mix-blend} training iteration: following \eqref{eq:all_mixup_learning}, the forward and background images randomly chosen from the training set are blended $M$ times, one per blending function approximator ($M=3$ in our implementation). Then, a weighted sum of all the blended images is performed with random weights $\boldsymbol{\lambda} = (\lambda_1, \lambda_2, \lambda_3)$ sampled from a Dirichlet distribution (see fig. $1$ in the supplementary material). As we model $f_{\boldsymbol{\theta}}$ as a neural network, this figure illustrates the way of adding a training instance into each iteration's mini-batch.
	}
	\label{fig:multi_blend_mix_blend}
\end{figure*}

The way of modelling $X$ in \eqref{eq:model_x} requires us to know the ideal blending function $\phi^{*}$.
As this is not the case, we opt to model blending in a probabilistic manner.
The composited labels $Y$ are trivially obtained by keeping the foreground labels $Y_F$ irrespective of the blending function (see fig. \ref{fig:modelling_of_x}).
Therefore, we define $X$ and $Y$ as:
\begin{equation}
\begin{split}
    X &= \Phi(X_F, X_B) \\
    Y &= Y_F
    \label{eq:generalized_x}
\end{split}
\end{equation}
where $\Phi$ is now a random variable whose observations $\phi$ are blending functions. Such modelling leads to the following joint probability measure:
\begin{equation}
\begin{split}
    \hat{P}_{X_F, X_B, Y, \Phi}&(\boldsymbol{x_f}, \boldsymbol{x_b}, \boldsymbol{y_f}, \boldsymbol{y_b}, \phi) = \\
    %\frac{1}{N_f N_b M}\sum_{i = 1}^{N_f}\sum_{j = 1}^{N_b}\sum_{m = 1}^{M}
    &\frac{1}{N_f N_b}\sum_{i, j}
    \delta_{\boldsymbol{x_f}_{i}}(\boldsymbol{x_f})
    \delta_{\boldsymbol{x_b}_{j}}(\boldsymbol{x_b})
    \delta_{\boldsymbol{y_f}_{i, j}}(\boldsymbol{y})
    P_{\Phi}(\phi)
    \label{eq:generalized_pm}
\end{split}
\end{equation}
where $i \in \{1, ..., N_f\}$, $j \in \{1, ..., N_b\}$, $N_f$ and $N_b$ are the number of foregrounds and backgrounds, and $P_{\Phi}$ represents the probability measure for $\Phi$. 

Given the framework presented in \eqref{eq:generalized_x} and \eqref{eq:generalized_pm}, we can now define $P_{\Phi}$ in a number of ways. 
%As we are unsure of the best approach to blend $\boldsymbol{x_f}$ onto $\boldsymbol{x_b}$ for learning segmentation, 
A possible approach is to arbitrarily choose a pool of blending methods to create our training images (as informally proposed in \cite{Dwibedi2017} for object detection), which is equivalent to defining $P_{\Phi}$ by:
\begin{equation}
    \label{eq:prob_of_phi}
    P_{\Phi}(\phi) = \frac{1}{M} \sum_{m} \delta_{\boldsymbol{\phi}_{m}}(\boldsymbol{\phi})
\end{equation}
where $m \in \{1, ..., M\}$, $M$ is the number of blending functions that we arbitrarily decide to define, and $\delta_{\boldsymbol{\phi}_m}$ stands for Dirac measures centered at $\phi_m$. This initial definition of $P_{\Phi}$ allows for either deterministic ($M=1$) or probabilistic ($M > 1$) blending, and turns our learning problem \eqref{eq:expected_risk_minimization} into:
\begin{equation}
\begin{split}
    \boldsymbol{\hat{\theta}} &= \underset{\boldsymbol{\theta}}{\argmin} %\mathop{{}\mathbb{E}}{}_{(F_G,B_G,Y,\Phi) \sim \hat{P}_{F_G,B_G,Y,\Phi}}
    \mathop{{}\mathbb{E}}{}_{\hat{P}_{X_F, X_B, Y,\Phi}}
    \left[ \ell \Big( f_{\boldsymbol\theta} \big( \Phi(X_F, X_B) \big), Y \Big) \right] \\
    %\boldsymbol{\hat{\theta}} 
    &= \underset{\boldsymbol{\theta}}{\argmin}
    %\frac{1}{N_f N_b M} 
    \sum_{i, j, m}
    \ell \Big( f_{\boldsymbol\theta} \big( \phi_m(\boldsymbol{x_f}_i, \boldsymbol{x_b}_j) \big) , \boldsymbol{y}_{ij} \Big)
\end{split}
    \label{eq:all_images_learning}
\end{equation}
The training scheme in \eqref{eq:all_images_learning}, which we denote as \textit{multi-blend}, is illustrated in fig. \ref{fig:multi_blend_mix_blend}, and formalizes the heuristic proposed experimentally by \cite{Dwibedi2017}.

In the case of $M=1$, all images in the training set are blended using the same method. If not chosen carefully, features of the blending could be erroneously learned during training, leading to poor generalization to real images.
%and erroneously identify the foreground
%will be formed by different pairs of ($\boldsymbol{x_f}$, $\boldsymbol{x_b}$) but 
 
Using several blending functions ($M > 1$) is a way to introduce robustness. Every pair  ($\boldsymbol{x_f}$, $\boldsymbol{x_b}$) added to the training set is blended $M$ times, and $M$ images are added to the training set. The intuition is that by having images whose only difference is the blending approach (as they have the same $\boldsymbol{x_f}$ and $\boldsymbol{x_b}$) we could potentially induce $f_{\boldsymbol{\theta}}$ to be \textit{blending invariant}.

Departing from the approach in \cite{Dwibedi2017} formalised above, rather than minimizing the risk functional defined only by a fixed set of $M$ blending functions, we now propose to delve into \textit{blending invariance} by modelling $X$ as:
\begin{equation}
    X = \sum_{m} \lambda_m \phi_m(X_F, X_B)
    \label{eq:weighted_sum}
\end{equation}
where $\lambda_m$ is the $m^{th}$ component of a vector of positive reals $\boldsymbol{\lambda}=(\lambda_1, ..., \lambda_M)$ s.t. $(\lambda_{m})_{m=1}^{M} \in \left]0, 1\right[^M$, $\sum_{m=1}^{M}\lambda_m = 1$, and $\{\phi_m\}_{m=1}^{M}$ is a basis of blending functions in our model. We model $\boldsymbol{\lambda}$ as an observation of a random variable $\Lambda \sim Dir(\boldsymbol{\alpha})$, parameterized by a vector of strictly positive reals $\boldsymbol{\alpha} = (\alpha_1, ..., \alpha_M)$ s.t.
\begin{equation}
    P_{\Lambda}(\boldsymbol{\lambda}) = \frac{1}{B(\boldsymbol{\alpha})} 
    \prod_{m} \lambda_{m}^{\alpha_m - 1}
    \label{eq:our_modelling_of_x_mix_blend}
\end{equation}
where $B(\boldsymbol{\alpha})$ is the multivariate beta function. This modelling of $X$ is a fundamental difference to \cite{Dwibedi2017}. It allows us to generate an infinite amount of images for a given pair foreground-background.
In contrast, in \cite{Dwibedi2017}, a particular combination of objects, which would be equivalent to our foreground-background pairs, can lead only to $M$ blended images (as many as blending functions employed). That is, we explore a wider space of blending functions.

Following our choice of probability measure for $P_{\Lambda}$, and hence for $\hat{P}_{X_F, X_B, Y, \Lambda}$, our learning problem \eqref{eq:expected_risk_minimization} turns into:
% This line with a space below is needed for the \small tag

% This empty line above is needed before \small
\small
\begin{equation} \label{eq:all_mixup_learning}
\begin{split}
    \boldsymbol{\hat{\theta}} &= \underset{\boldsymbol{\theta}}{\argmin}
    \mathop{{}\mathbb{E}}{}_{\hat{P}_{X_F, X_B, Y, \Lambda}}
    \left[
    \ell \left[ f_{\boldsymbol\theta} 
    \left( 
    \sum_{m}
    \Lambda_m \phi_m\left(X_F, X_B\right) 
    \right), Y 
    \right]
    \right]
    \\
    &= \underset{\boldsymbol{\theta}}{\argmin}
    \sum_{i, j}   
    \int_{\boldsymbol{\lambda}}
    \ell \left[ f_{\boldsymbol\theta} \left( 
    \sum_{m} 
    \lambda_m \phi_m(\boldsymbol{x_f}_i,\boldsymbol{x_b}_j) \right),\boldsymbol{y}_{ij} \right]
    P_{\Lambda}(\boldsymbol{\lambda})d\boldsymbol{\lambda}
\end{split}
\end{equation}
\normalsize
where $\Lambda = [\Lambda_1, ..., \Lambda_M]$. 
The training strategy presented in \eqref{eq:all_mixup_learning} requires the computation of the loss over all the combinations of foreground-background $(\sum_{i, j})$ for all the possible weighted sums ($\int_{\boldsymbol{\lambda}}$) of blended images. 
Although this is unfeasible, in practice, foregrounds, backgrounds, and weights $(\boldsymbol{\lambda})$ can be (and are) randomly sampled during the training of the network with SGD. In that case the network training process is therefore solving the following optimization problem:
\begin{equation} \tag{12}
\boldsymbol{\hat{\theta}} = \underset{\boldsymbol{\theta}}{\arg\min}
\sum_{i, j, \upsilon}   
\ell \left[ f_{\boldsymbol\theta} \left( \sum_{m} \lambda_{m, \upsilon} \phi_m(\boldsymbol{x_f}_i, \boldsymbol{x_b}_j) \right), \boldsymbol{y}_{ij} \right]
\label{eq:all_mixup_learning_sampled}
\end{equation}
where $\upsilon \in \{1, ..., \Upsilon\}$, $\Upsilon$ is the number of samples of $\boldsymbol{\lambda}$ drawn according $P_{\Lambda}(\boldsymbol{\lambda})$ during training for each combination foreground-background, and $\ell$ represents the pixel-wise cross-entropy loss employed by our segmentation network during training. $\Upsilon$ is proportional to the number of optimization steps or training iterations selected. We refer to the learning strategy in \eqref{eq:all_mixup_learning_sampled} as \textit{mix-blend} learning (see fig. \ref{fig:multi_blend_mix_blend} and supplementary material fig. $1$).

%\subsection{Unsupervised domain adaptation}
\subsection{Post-processing}
\label{unsupervised_domain_adaptation}
In our method, the foreground images $\boldsymbol{x_f}$ are recorded in
a loosely controlled  and somewhat artefactual setup.
%the wild.
The illumination conditions (LED light source), recording devices (mobile phone and DSLR), and camera viewpoints to record the instruments are different from those seen in real clinical videos. In addition, we just recorded a small sample of instruments (see supplementary material fig. $4$). % not all possible instruments that can be found in clinical videos.
%There are the ones present in the clinical testing set.
One could argue that mimicking the clinical setup by recording with different endoscopes, a laparoscopic phantom, and a large number of surgical instruments could lead to more realistic blended images and better performance. However, there is no guarantee, that after creating such setup, a domain gap would not still exist between semi-synthetic data and real clinical videos. What is guaranteed is that the method would be less flexible.
Hence, we opt to mitigate the expected domain gap between our trivial setup and real clinical videos with post-processing.

%Also there are many types of endoscopes, if the one we use to record the instruments has some optical characteristics different from other endoscopes we will be in a similar situation as if we just use a mobile phone camera, we are going to need domain adaptation techniques.

GrabCut \cite{Rother2004a} is a well-known semi-automatic segmentation technique that may be employed for post-processing (without needing to provide manual scribbles). The probability map generated by our neural network may replace the usual manual \textit{scribble} that is employed to initialize the Gaussian mixture models of GrabCut. We assume that network estimated probabilities $<0.2$ represent \textit{sure background}, and $\geq0.8$ \textit{sure foreground}. %The \textit{sure} pixels transforms into an infinite capacity for the arcs that connect those pixels with the foreground and background graph nodes. As a consequence, 
The segmentation of pixels for which the network prediction is considered certain is not modified by GrabCut.
Provided the seeding process proves reliable,
by expanding the segmentation from these certainty zones according to colour contrast present in the specific image, GrabCut enables to bridge relatively minor domain gaps.
\section{Materials}
\label{Materials}
In order to implement the training schemes in \eqref{eq:all_images_learning} and \eqref{eq:all_mixup_learning}, we need to devise:
\begin{enumerate}
    \item A way to obtain \textit{source images}, where \textit{source images} denotes images that are easy to segment into $\boldsymbol{x_f}$ and $\boldsymbol{x_b}$.
    \item A set of blending methods $\{\phi_m\}_{m=1}^{M}$ that allows us to combine $\boldsymbol{x_f}$ and $\boldsymbol{x_b}$ to form new images.
\end{enumerate}
In this work, we propose to obtain two types of \textit{source images}, foregrounds and backgrounds.

\subsection{Background dataset collection}
Although recording videos containing just tissue may be possible prospectively, for this work and without loss of generality,  we have obtained all backgrounds from freely available surgical procedures on the Internet, as done in other computer vision datasets \cite{Lin2014}. %(list of URLs in section \ref{appendix_videos} of supplementary material). 
We manually select frames that only contain human tissue from video sequences of different surgical procedures. %(typically from the beginning of the sequence, when instruments have not been inserted yet).
Segmenting these background \textit{source images} into components is trivial. There is no foreground component in them, just background tissue. Hence, they represent our $\boldsymbol{x_b}.$
We have collected $6130$ images from $50$ laparoscopic interventions (exemplar images illustrated in fig. $5$ of the supplementary material).
%
%containing artifacts such as remainings of tool-tissue interactions, smoke, blood, debris, and shadows 
%
%
The background images included in this dataset contain some degree of tool-tissue interaction. They display direct interactions, such as tissue being pulled (with the tools out of the camera view), and also indirect artefacts, such as those inflicted by the instruments on the background tissue. Examples of the latter are shadows, blood, debris, and smoke. Nonetheless, it should be noted that there may be a certain type of interaction between instruments and tissue that we may not be able to reproduce via image compositing. This may impact the generalization ability of the model. However, as our clinical testing set (RoboTool) contains these interactions, the results shown in section VII already account for the impact of this limitation. A possible workaround is to add a small amount of manually annotated images displaying special tool-tissue interactions that cannot be observed otherwise. Once we have the tools segmented out from those images, we can refill the tool pixels with other background images as we do for our flying distractors. Then, these backgrounds become fully functional as those that do not contain any tools.
The background images used to build the semi-synthetic training dataset are not present in any of the testing sets employed.

\subsection{Foreground dataset collection}
To extract foreground components, we collect a pool of instruments and place them over a \textit{chroma key} (see fig. $2$ and $3$ in the supplementary material). These images represent our second type of \textit{source images}.
In this scenario, we can reproduce many different lighting conditions and viewpoints. As the chroma key is monochromatic (green), we can automatically segment $\boldsymbol{x_f}$ (tools), and discard the green background component.
We have recorded two subsets, each one with a different camera. To facilitate segmentation, 
the chroma key has to be properly illuminated. 
This way, the amount of shadows on the chroma fabric is minimized. 
The number of instruments per image varies from one to three (out of a total of $17$ instruments recorded). The two recording devices employed are a mobile phone camera, whose subset contains $13613$ frames of size $4032 \times 3024$ pixels that display a single instrument over the chroma key, and a DSLR camera, whose subset contains $567$ frames of size $3360 \times 2240$ pixels.

Although being able to record foregrounds with a commercial phone or a DSLR adds flexibility to the method, a possible approach to reduce the domain gap between synthetic and real data could be to record the foregrounds with the target imaging system. This requires having access to the exact imaging device used in practice, and generalization to other make and models (and their evolution in time) may still be limited. Nonetheless, it would be interesting for future work to examine the generalization performance when recording with different endoscopes.

\subsection{Foreground dataset segmentation}
\label{subsection_foreground_dataset_segmentation}
Images are converted to HSV and thresholded to capture the green pixels that belong to the background. 
The mask generated by the HSV threshold is provided as a unitary term prior to \textit{GrabCut} \cite{Rother2004a}. 
These automatic segmentations are quality controlled by means of visual inspection. Those few with obvious inaccuracies due to a GrabCut failure (e.g. green area captured as tool or instrument missing parts) are excluded.

As thresholding is such a simple technique, different levels of lighting affect the quality of the results significantly. It is convenient to tune the HSV threshold right before recording
so that it can be adjusted to the lighting conditions of the scene and avoid tedious postprocessing.
To reduce noise in the automatically-generated tool segmentation masks, our interface allows specifying the number of instruments being recorded so only those HSV-thresholded pixels that lie inside the largest $N_i$ connected components are kept, where $N_i$ is the number of instruments.

\subsection{Semi-synthetic dataset: training and validation}
Although our image synthesis method may be performed on-the-fly, we precompute $100$K images blended with all our basis (see section \ref{blending_strategy}) to speed up our semi-synthetic training. Only the Dirichlet random weighted sum (eq. \ref{eq:weighted_sum}) is performed on-the-fly. For validation, we precompute a small semi-synthetic dataset of $500$ images that use $392$ foregrounds recorded over a red chroma key, and $428$ backgrounds that were kept aside from the background dataset. This small semi-synthetic dataset is just used as a baseline for early stopping. That is, as stopping criteria for the training on semi-synthetic data.

%%%%%%%%%%%%%%%%%%%%%%%%%%%%%%%%%%%%%%%%%%%%%%%%%%%%%%%%%%%%%
\subsection{EndoVis2017 (existing real pre-clinical dataset): testing set}
\label{sec:testing_set_endovis}
%%%%%%%%%%%%%%%%%%%%%%%%%%%%%%%%%%%%%%%%%%%%%%%%%%%%%%%%%%%%%
For evaluation, we use the images coming from the training set given at the 2017 Robotic Instrument Segmentation Challenge \cite{Allan2019}. 
As these images come from recordings made with the da Vinci Surgical System (dVSS) and have been manually labeled, we refer to them as real data (as opposed to semi-synthetic data, which is the one we generate with our method). 
We use the training set of the challenge for evaluation (annotations are widely available).
This dataset comes with eight video sequences. In order to generate baseline results for comparison with our method, we use the same protocol of the challenge. We perform cross-validation with eight folds. In each fold, the testing set contains only one video. The remaining seven videos are used for training and validation ($10\%$ of the video frames in these seven videos are left for early stopping, $90\%$ for training).
The only cross-validation performed during our experiments is that mentioned in this section, aimed at evaluating the baseline performance on the EndosVis2017 dataset.

%%%%%%%%%%%%%%%%%%%%%%%%%%%%%%%%%%%%%%%%%%%%%%%%%%%%%%%%%%%%%%
\subsection{RoboTool (new real clinical dataset): testing set}
\label{sec:testing_set_robotool}
%%%%%%%%%%%%%%%%%%%%%%%%%%%%%%%%%%%%%%%%%%%%%%%%%%%%%%%%%%%%%%
We make public our newly created real clinical testing set called RoboTool (see fig. $6$ of the supplementary material), containing $514$ manually annotated images extracted from the videos of $20$ freely available surgical procedures. For those baseline experiments where a network is trained on RoboTool, the validation set used for early stopping consists of $51$ images that have not been seen in training and come from other surgical procedures different from the $20$ employed to build this dataset.

\section{Implementation of the methods}
\label{section_implementation}
%All semi-synthetic data and code corresponding to the implementation of the methods 
%detailed in this section 
%will be released open-source upon publication of this manuscript.
All  semi-synthetic   data  and  code   corresponding  to  the implementation  of  the  methods is made available in open access\footnote{\url{https://synapse.org/synthetic}}.

\subsection{Data augmentation and standardization}
\label{preprocessing_section}
After obtaining images for the foreground (based on chroma key) and background (from the Internet),
and prior to their superimposition, we augment and standardize them as detailed in the following paragraphs. This step is not detailed in our mathematical model in section \ref{section_methods} for the sake of conciseness, although the extension of the model to account for it is trivial.

We perform different augmentations on foreground, background, and blended images. Foreground tools are randomly zoomed, rotated, and vertically and horizontally flipped and shifted. All these operations are performed while keeping the tools connected to the border of the image. In addition, foregrounds have synthetic blood droplets and tissue debris blended onto the tools (see fig. $7$ of the supplementary material). Their brightness is also randomly altered.
%
%and artificial blood droplets blended onto them. 
Background augmentations comprise horizontal and vertical flips, brightness changes, and random rotations of $90$ degrees. Blended images are augmented with a set of techniques from Albumentations~\cite{Buslaev2020}. Namely, cutouts, synthetic smoke and shadows, JPEG compression, RGB and HSV shifts, noise (multiplicative, Gaussian, ISO), and blur (Gaussian, motion, median). In addition to these, backgrounds are also augmented with \textit{flying distractors} and endoscopic padding. \textit{Flying distractors} are cutouts of other backgrounds blended with the shape of a random foreground tool. For any given training image, the blending function used to superimpose the foreground tools is also used to blend the \textit{flying distractors}. Endoscopic padding consists of simulating the black border occasionally present in endoscopic images. We randomly pad the images enclosing the frame with a rectangular or circular shape. Gaussian noise is randomly added to the black padding. A set of exemplary semi-synthetic images is shown in fig. $7$ of the supplementary material.

Prior to the blending, the augmented pairs $(\boldsymbol{x_f}$, $\boldsymbol{x_b})$ are resized to our standardized width, $640$-pixel (i.e. original aspect ratio is kept). A random crop is performed on the element of the pair of larger height so that both display the same height.
After this step, both $\boldsymbol{x_f}$ and $\boldsymbol{x_b}$ have the same resolution. %Furthermore, the instruments present in the $\boldsymbol{x_f}$ image are correctly located within the image, stemming from the border, as in most real images. 
This facilitates the blending of tools onto tissue ($\boldsymbol{x_f}$ over $\boldsymbol{x_b}$), described below.

\subsection{Blending}
\label{blending_strategy}
In equations \eqref{eq:all_images_learning} and \eqref{eq:all_mixup_learning}, we defined two ways of learning $\boldsymbol{\theta}$ 
for our instrument-background segmentation function $f_{\boldsymbol{\theta}}$.
%depending on how the blending is performed. 
Both approaches rely on the existence of a set of blending functions $\{\phi_m\}_{m=1}^{M}$ that we can evaluate to obtain a training image from a pair of $\boldsymbol{x_f}$ and $\boldsymbol{x_b}$. In our implementation, we evaluate these functions using $M=3$ blending or superimposition algorithms:
\begin{itemize}
    \item Trivial blending. The pixels activated in the tool mask are copied from $\boldsymbol{x_f}$ onto $\boldsymbol{x_b}$ to form the final blend $\boldsymbol{I_{N}}$.
    \item Gaussian feathering. The foreground segmentation mask $\boldsymbol{m}$ is eroded ($k=3$) and blurred ($k=5$). The final image is generated as
    %
    %\begin{equation}
    $\boldsymbol{I_{G}} = \boldsymbol{m} \cdot \boldsymbol{x_f} + (1 - \boldsymbol{m}) \cdot \boldsymbol{x_b}$,
    %\end{equation}
    %
    where $\boldsymbol{m}$ represents the mask after erosion and blurring.
    \item Laplacian pyramid blending \cite{Burt1983}. A Laplacian pyramid is constructed for both images. A Gaussian pyramid is built for the region occupied by $\boldsymbol{m}$. Then, Laplacian pyramids are combined using the nodes of the Gaussian pyramid as weights and collapsed to form the blended image $\boldsymbol{I_{L}}$. 
\end{itemize}
%
%(see section \ref{blending_strategy} for a detailed explanation of how these blending functions are used):
%depending on the blending strategy employed (several presented in section ), we :
%
%See section \ref{blending_strategy} for a detailed explanation on how these blending algorithms are employed during training.
%
Given these blending basis, our implementation of \textit{multi-blend} learning \eqref{eq:all_images_learning} consists of populating our training set with three images (one per blending method) for each pair $(\boldsymbol{x_f}, \boldsymbol{x_b})$. %In our experiments we create datasets of 1K, 2K, 5K, and 10K pairs (see table \ref{tab:dataset_splits}).

For our implementation of \textit{mix-blend} \eqref{eq:all_mixup_learning} we choose the same (to be able to compare results) combinations of $(\boldsymbol{x_f}, \boldsymbol{x_b})$ selected for the experiments of \eqref{eq:all_images_learning}.
%blending them $M=3$ times, one per algorithm, as done for \textit{multi-blend}. 
%as proposed in~\eqref{eq:weighted_sum}, and following the integral in \eqref{eq:all_mixup_learning}, 
However, in contrast to \eqref{eq:all_images_learning}, when the optimization problem \eqref{eq:all_mixup_learning} is solved with SGD, the training samples included in each mini-batch are generated on the fly. To generate each sample we select a pair of $(\boldsymbol{x_f}, \boldsymbol{x_b})$, blend it $M=3$ times, draw a random sample $\boldsymbol{\lambda}$ (vector of three weights, one per blending method) from $Dir(\boldsymbol{\alpha})$ with $\boldsymbol{\alpha} = (1.0, 1.0, 1.0)$, and perform a weighted sum of the $M=3$ blended images.

\subsection{Network architecture and training protocol}
\label{network_architecture}
%As researching network architectures is not within the scope of this article, and 
As the leading approach of the $2017$ Robotic Instrument Segmentation Challenge \cite{Allan2019} was a U-Net %($32$ slices in the first layer) 
\cite{Ronneberger2015}, this encoder-decoder architecture was chosen to model our instrument segmentation function $f_{\boldsymbol{\theta}}$. 

%\subsubsection{Training protocol}
%^\label{training_protocol}
We train all the networks using the same protocol. A fixed learning rate (LR) of $0.001$. 
%and the baseline (trained on real images) with $1e-2$ ($1e-2$ delivers superior results on validation compared to $1e-3$ and $1e-4$). We decay the LR every $10$ epochs by $0.5$. 
A batch size of $32$ because it is the maximum our GPU can fit.
%and resize all the images (respecting the form factor) to a $512$-pixel width for the baseline network. %The semi-synthetic images are already generated at this dimension. 
Early stopping (ES) is used to bound the duration of our training. The ES baseline is always the validation set of each experiment. The minimum delta is set to $0.01$ of absolute average mIoU and the patience to $20$ epochs.
\rebnew{
All our networks are trained with SGD with momentum $0.9$ and the widely used pixel-wise cross-entropy as loss function:
\begin{equation}
\label{eq:evaluation_metric}
\mathcal{L}(\boldsymbol{\hat{y}}, \boldsymbol{y}) =
- \sum_{i=1}^{N_p} \sum_{k=1}^{K} y_{i,k} \log{\hat{y}_{i,k}}
\end{equation}
where $\boldsymbol{\hat{y}}$ is the predicted segmentation label, $\boldsymbol{y}$ is the ground truth label, $i \in \{1,...,N_p\}$ where $N_p$ is the number of pixels, and $k \in \{1, ..., K\}$ where $K=2$ is the number of classes.
}

\section{Evaluation}\label{section_evaluation}
\rebmod{%
Although metrics such as the Frechet Inception Distance (FID)~\cite{Heusel2017} could be useful to evaluate the similarity between the generated and real data, in this work, we are not aiming to create photo-realistic images, but rather to train a network that generalizes well to real data for the tool segmentation task. 
In fact, the generated semi-synthetic images contain \textit{flying distractors} (see section \ref{preprocessing_section}), which are cutouts blended with the shape of a tool and the texture of a background. These artefacts are not present in real images, but they help to learn the segmentation by encouraging the network to not learn the blending as a feature to detect tools. Therefore, evaluating the realism of the semi-synthetic images would not lead to a meaningful result. Our evaluation focuses on assessing the quality of the tool segmentation. For such purpose, we employ the widely used intersection over union (IoU), also called Jaccard index.
}
%Our evaluation metric is the widely used intersection over union (IoU), also called Jaccard index. 
%
For a single video frame, we compute the IoU $\mathcal{J}$ between the binarized (threshold $\geq0.5$, background = 0, tool = 1) probability prediction $b(\mathbf{\hat{y}})$ and the ground truth $\mathbf{y}$ (which is already a binary image). That is:
\begin{equation}
\label{eq:evaluation_metric}
\begin{split}
\mathcal{J}(b(\mathbf{\hat{y}}), \mathbf{y}) &=
\frac
% Numerator:
{\sum_{i=1}^{N_p}\hat{y}_{i,k} \cdot y_{i, k} + \epsilon}
% Denominator:
{\sum_{i=1}^{N_p}\hat{y}_{i,k} + \sum_{i=1}^{N_p} y_{i, k} + \epsilon}
\end{split}
\end{equation}
where $b$ denotes the binarization function, $N_p$ is the number of pixels in the image, $\epsilon$ is the machine epsilon, and $\mathcal{J}$ bounds our scores to the interval $[0, 1]$. To report the IoU for a video sequence, we average the metric $\mathcal{J}$ across all the frames. All the results are given in percentage.
\section{Results and Discussion}
\begin{figure}[t!]
	\includegraphics[width=\columnwidth]{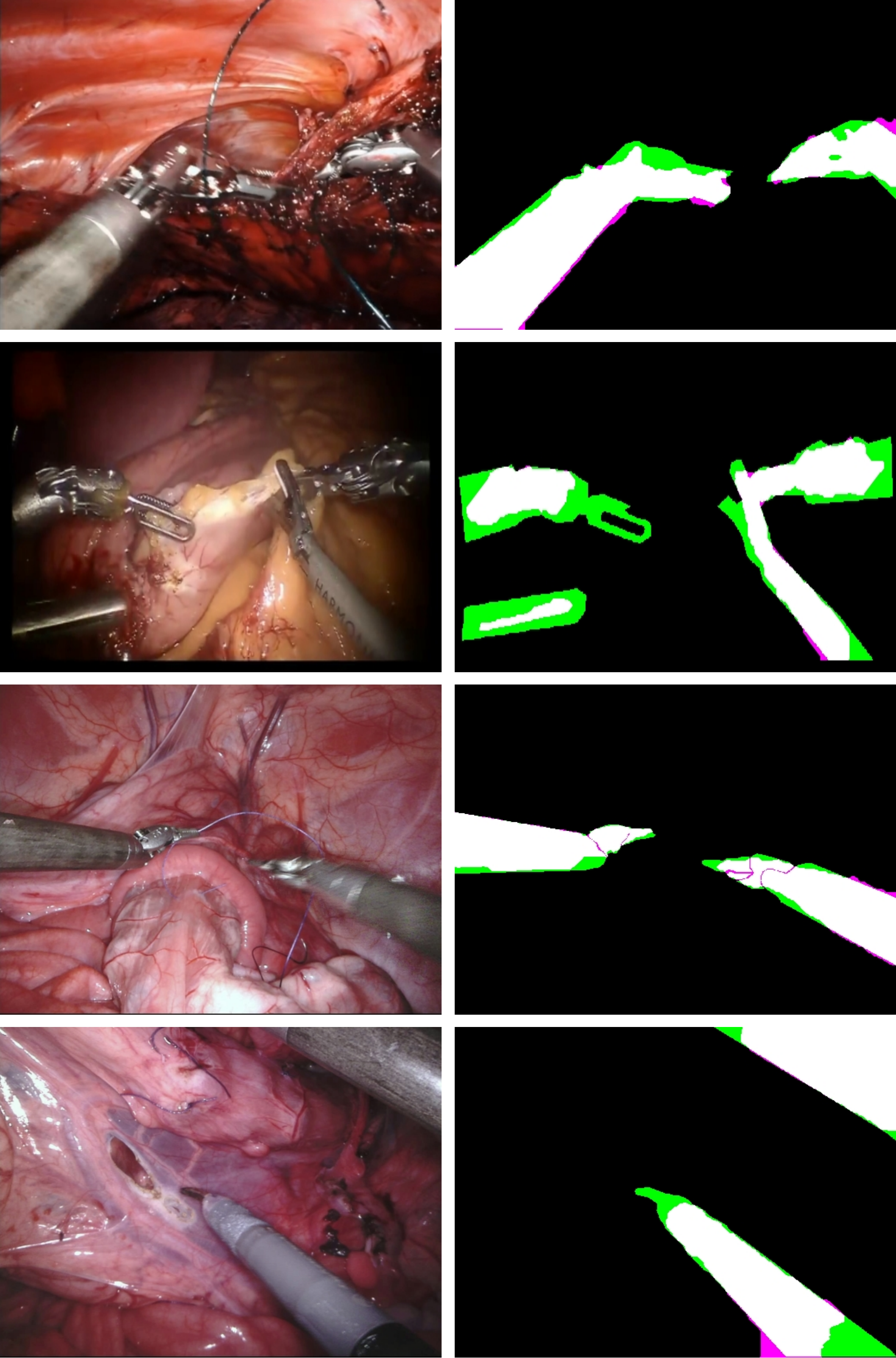}
    \caption{Exemplary segmentations of the RoboTool dataset (two top rows) and EndoVis2017 dataset (two bottom rows) segmented with our \textit{mix-blend} method. The best/median/worst cases for each testing dataset are shown in figures $8$-$10$ of the supplementary material.
    %+ Fourier (eq. \ref{eq:fourier_da}) and GrabCut domain adaptation. 
    Confusion images are displayed on the right column. True positive (white), true negative (black), false positive (magenta), and false negative (green).}
	\label{fig:qualitative_results}
\end{figure}
\begin{table*}[!htbp]
	\centering
	\caption{Baseline results. Training and evaluating on manually labeled real data. No semi-synthetic data nor blending technique has been used to generate the results on this table. In the first line of the results, when training and testing on EndoVis2017, leave one out is used, following the protocol of the EndoVis2017 challenge (see section \ref{sec:testing_set_endovis}). 
	%EM stands for entropy minimization. The dataset used for domain adaptation is a third dataset (see section \ref{unlabelled_dataset}), independent from EndoVis2017 and RoboTool.
	}
	\begin{tabular}{lccc}
		\hline
		\multicolumn{1}{c}{\bfseries Training dataset} &
		\multicolumn{1}{c}{\bfseries Post-processing} &
		\multicolumn{1}{c}{\bfseries Testing dataset} &
		\multicolumn{1}{c}{\bfseries IoU [5\%, 95\%]} \\
		\hline
		EndoVis2017  &  None                                            &  EndoVis2017  &  $81.6~[69.7, 89.6]$  \\ % Exp: [144, 151], leave one out
	    \hline
	    RoboTool     &  None                                            &  EndoVis2017             &  $73.8~[56.3, 86.6]$  \\ % Exp: 157
	    %RoboTool     &  Fourier (eq. \ref{eq:fourier_da})               &  EndoVis2017  &  $74.5~[54.7, 85.6]$  \\ % Exp: 157
	    RoboTool     &  GrabCut                                         &  EndoVis2017             &  $80.5~[55.8, 94.1]$  \\ % Exp: 157
	    %RoboTool     &  Fourier (eq. \ref{eq:fourier_da})    + GrabCut  &               &  $79.9~[51.1, 94.4]$  \\ % Exp: 157
	    \hline
	    EndoVis2017  &  None                                            &  RoboTool             &  $66.6~[43.0, 87.2]$  \\ % Exp: 152
 		%EndoVis2017  &  Fourier (eq. \ref{eq:fourier_da})               &  RoboTool     &  $66.1~[40.9, 87.5]$  \\ % Exp: 152
 		EndoVis2017  &  GrabCut                                         &  RoboTool             &  $69.4~[37.7, 91.8]$  \\ % Exp: 152
 		%EndoVis2017  &  Fourier (eq. \ref{eq:fourier_da}) + GrabCut     &               &  $67.4~[34.1, 91.4]$  \\ % Exp: 152
	\end{tabular}
	\vspace{0.2cm}
	\label{tab:real_results}
\end{table*}

\begin{table*}[!htbp]
	\centering
	\caption{Ablation study of blending methods. Training on semi-synthetic data and testing on unseen real data.} %No domain adaptation techniques used.}
	\begin{tabular}{lcc}
		\hline
		\multicolumn{1}{c}{\bfseries Training dataset (blending)} &
		\multicolumn{1}{c}{\bfseries Testing dataset} &
		%\multicolumn{1}{c}{\bfseries Training mIoU [5\%, 95\%]} & 
		\multicolumn{1}{c}{\bfseries IoU [5\%, 95\%]} \\
		\hline 
		Semi-synthetic (Trivial)                         &               &  $53.7~[44.3, 66.6]$           \\ % Exp: 153
		Semi-synthetic (Gaussian)                        &               &  $55.2~[44.8, 70.4]$           \\ % Exp: 154
		Semi-synthetic (Laplacian)                       &  EndoVis2017  &  $68.3~[52.4, 83.1]$           \\ % Exp: 155
		Semi-synthetic (Multi-blend) \cite{Dwibedi2017}  &               &  $64.3~[49.2, 79.8]$           \\ % Exp: 156
		Semi-synthetic (Mix-blend)                       &               &  $\textbf{72.8}~[56.5, 87.8]$  \\ % Exp: 138
		\hline
		Semi-synthetic (Trivial)                         &               &  $48.4~[38.2, 65.5]$           \\ % Exp: 153		
		Semi-synthetic (Gaussian)                        &               &  $48.7~[38.2, 65.7]$           \\ % Exp: 154
		Semi-synthetic (Laplacian)                       &  RoboTool     &  $54.3~[40.4, 75.6]$           \\ % Exp: 155
		Semi-synthetic (Multi-blend) \cite{Dwibedi2017}  &               &  $51.7~[39.5, 74.5]$           \\ % Exp: 156
		Semi-synthetic (Mix-blend)                       &               &  $\textbf{56.1}~[40.2, 77.5]$  \\ % Exp: 138
	\end{tabular}
	\vspace{0.2cm}
	\label{tab:blending_results}
\end{table*}

\begin{table*}[!htbp]
	\centering
	\caption{Results of our proposed blending method in combination with post-processing.}
	\begin{tabular}{lccc}
		\hline
		\multicolumn{1}{c}{\bfseries Training dataset} &
		\multicolumn{1}{c}{\bfseries Post-processing} &
		\multicolumn{1}{c}{\bfseries Testing dataset} &
		\multicolumn{1}{c}{\bfseries IoU [5\%, 95\%]} \\
		\hline
		%                            & Fourier (eq. \ref{eq:fourier_da})            &               &  $74.0~[58.7, 87.5]$           \\ % Exp: 138
		Semi-synthetic (Mix-blend)  & GrabCut                                      &  EndoVis2017  &  $83.3~[62.7, 93.9]$  \\ % Exp: 138
		%                            & Fourier (eq. \ref{eq:fourier_da}) + GrabCut  &               &  $82.4~[61.3, 93.8]$           \\ % Exp: 138
		\hline
		 %                           & Fourier (eq. \ref{eq:fourier_da})            &               &  $60.2~[42.4, 80.5]$           \\ % Exp: 138
		 Semi-synthetic (Mix-blend) & GrabCut                                      &  RoboTool     &  $68.1~[42.6, 92.5]$           \\ % Exp: 138
		 %                           & Fourier (eq. \ref{eq:fourier_da}) + GrabCut  &               &  $\textbf{68.9}~[43.7, 91.4]$  \\ % Exp: 138
	\end{tabular}
	\vspace{0.2cm}
	\label{tab:uda_results}
\end{table*}

In our first experiment, we train eight networks. Each one is trained on seven videos of the EndoVis2017 dataset, and tested on the remaining video (see results per video in table I of the supplementary material). This experiment leads to an average mIoU across experiments of $81.6~[69.7, 89.6]$ (confidence interval $[5\%, 95\%]$). At first sight, it could seem as if the binary segmentation of surgical tools is a solved problem. However, when we test a network trained on all the EndoVis2017 videos on RoboTool (our real clinical dataset presented in section \ref{sec:testing_set_robotool}), the performance drops to $66.6~[43.0, 87.2]$, suggesting some overfitting to the recording conditions of the challenge dataset. 
For a fair comparison with the proposed method, we apply post-processing to the output of the network trained on EndoVis2017 (table I), pushing the average mIoU on RoboTool to $69.4~[37.7, 91.8]$.
We also performed the inverse experiment, training on RoboTool and testing on EndoVis2017. This led to an average mIoU of $73.8~[56.3, 86.6]$. These results suggest that networks trained on these small manually labeled datasets (coming from a small number of recorded interventions) do not generalize as well as it could be expected. All the results for training and testing on real data are presented in table \ref{tab:real_results}. 
%
%We analyze the performance of each blending method individually, and compare the different approaches to combine them them. For doing this, we train on semi-synthetic data, and test on the two real datasets, EndoVis2017 and RoboTool. We carry out this comparison by training networks with identical $\boldsymbol{x_f}$ and $\boldsymbol{x_b}$ while changing the blending method. The results of our proposed \textit{mix-blend} learning are top performing (see table \ref{tab:blending_results}), 
%%(see predictions in fig. \ref{fig:mixup_results} and \ref{fig:mixup_pp_results}), 
%suggesting that varying the blending method helps to boost segmentation accuracy when jumping from semi-synthetic to real data. This effect has also been observed by \cite{Dwibedi2017}, whose \textit{multi-blend} implementation is actually a particular case of \textit{mix-blend}, say $\boldsymbol{\alpha} = (0.001, 0.001, 0.001)$.  
%%learning approach we introduce in \eqref{eq:all_images_learning}. 
%We believe the reason why \textit{mix-blend} learning -- eq. \eqref{eq:all_mixup_learning} -- achieves higher IoU than \textit{multi-blend} -- eq. \eqref{eq:all_images_learning} -- is because it explores a larger (infinite) variety of possible blendings (not just the $M$ basis), delving deep into the invariance to the blending mechanism.

\rebmod{%
In the context of generating a dataset that can allow for the learning of the tool segmentation, crisp borders induced by simple copy-pasting represent a spurious feature that the network would exploit as a mean to solve for the segmentation of the tools in semi-synthetic data. To address this challenge, we analyze the performance of each blending method individually, and compare the different approaches to combine them.}
%We analyze the performance of each blending method individually, and compare the different approaches to combine them. 
%
For doing this, we train on semi-synthetic data, and test on the two real datasets, EndoVis2017 and RoboTool. We carry out this comparison by training networks with identical $\mathbf{x_f}$ and $\mathbf{x_b}$ while changing the blending method. 
Our results indicate that Laplacian blending is superior to both trivial and Gaussian blending. Surprisingly, it also outperforms \textit{multi-blend} by four percentage points, suggesting that the inclusion of either trivial, Gaussian, or both blending modes is counterproductive. In contrast, \textit{mix-blend} outperforms Laplacian by $4$ percentage points and \textit{multi-blend} by $8$ percentage points. This result supports our theoretical claim that \textit{multi-blend} is just a particular corner case of \textit{mix-blend} with $\mathbf{\alpha} = (0.001, 0.001, 0.001)$.
The top performing results of our proposed \textit{mix-blend} learning (see table II) also suggest that varying the blending method helps to boost segmentation accuracy when jumping from semi-synthetic to real data. This effect has also been observed by Dwibedi et al. in \cite{Dwibedi2017}.
We believe the reason why \textit{mix-blend} learning -- eq. $12$ -- achieves higher IoU than \textit{multi-blend} -- eq. $8$ -- is because it explores a larger (infinite) variety of possible blendings (not just the $M$ basis), delving deep into the invariance to the blending mechanism.

% BEFORE:
%
%The last part of our study is on bridging the gap between synthetic and real data. We show that by using two simple unsupervised domain adaptation methods, we are able to push the performance of our semi-synthetic \textit{mix-blend} method to reach the same accuracy as a network trained on real data. Training on EndoVis2017 and testing on RoboTool (our real clinical dataset with $20$ surgical procedures), we achieve $66.6[43.0, 87.2]$.
%
% AFTER: to accommodate for the reviewers' suggestion of not mentioning "domain adaptation" (which is now moved to the supplementary material based on their recommendations) but just stick to "post-processing" as suggested
%
The last part of our study is on bridging the gap between synthetic and real data. We show that by using simple post-processing, we are able to push the performance of our semi-synthetic \textit{mix-blend} method to reach the same accuracy as a network trained on real data. Training on EndoVis2017 and testing on RoboTool (our real clinical dataset with $20$ surgical procedures) we achieve $66.6~[43.0, 87.2]$. With GrabCut post-processing this increases to $69.4~[37.7, 91.8]$.
%
% BEFORE:
%
%Training on semi-synthetic data with \textit{mix-blend}, and applying domain adaptation we reach $68.9~[43.7, 91.4]$ on RoboTool (see exemplary segmentations in fig.~4). All the results of our complete pipeline trained on semi-synthetic data and evaluated on real data are shown in table \ref{tab:uda_results}.
%
% AFTER: to accommodate for the reviewers' suggestion of not mentioning "domain adaptation" (which is now moved to the supplementary material based on their recommendations) but just stick to "post-processing" as suggested
%
Training on semi-synthetic data with \textit{mix-blend}, we achieve $56.1~[40.2, 77.5]$ on RoboTool. With GrabCut post-processing we reach $68.1~[42.6, 92.5]$ (see exemplary segmentations in fig.~4). All the results of our complete pipeline trained on semi-synthetic data and evaluated on real data are shown in table \ref{tab:uda_results}.
Figures $8$, $9$, and $10$ of the supplementary material facilitate the visual comparison of results at several percentile levels for the different methods (best/median/worst cases). In fig. $8$ of the supplementary material, we show the baseline results (training on a real dataset, and testing on a different real dataset). In fig. $9$ and $10$ of the supplementary material, we show the best/median/worst cases when training on semi-synthetic data and testing on real datasets RoboTool and EndoVis2017.

\section{Conclusion}
We have shown a new method to automatically generate labels for surgical tool segmentation. Synthetically generating the whole surgical scene is a very challenging problem. However, just performing a simple semi-synthetic blending that explores the %vicinity
mix of a set of blending basis, and applying post-processing, we are able to train a convolutional neural network that achieves an analogous performance to that of a network trained on currently available manually labeled datasets such as EndoVis2017. %We have introduced a new method to explore the vicinity of blendings %inspired by the \textit{mixup} empirical risk minimization 
%that is able to match the segmentation performance of a U-Net trained on real data. F
Future work includes the exploration of domain adaptation techniques that could potentially push further the results obtained by the semi-synthetic blending. 
\FloatBarrier
\bibliography{library}
\bibliographystyle{IEEEtran.bst}

\clearpage

\appendix  % for no appendix heading
% do not use \section anymore after \appendix, only \section*
% is possibly needed

\section*{A. Fourier-based domain adaptation}
\label{sec:appendix_fourier}
To mitigate the expected domain gap between our trivial setup and real clinical videos we tested an approach derived from one of the top-performing \cite{Toldo2020} unsupervised domain adaptation methods proposed by Yang. et al in \cite{Yang2020}. As the performance boost achieved was not considerable, we decided to keep both the proposed method as well as the results achieved in the supplementary material. 

%The method proposed in \cite{Yang2020} is based on the Fourier transform.
%This transform has been employed for image processing for more than four decades \cite{Gonzalez1977}.
%Recently, Yang. et al \cite{Yang2020} have proposed to use it for unsupervised domain adaptation.
%As a zero-shot alignment of low level statistics, showing 
In \cite{Yang2020}, 
%which shows a state-of-the-art performance, 
source synthetic images are transformed to make them look closer to target real images. 
%This is done by They propose to map a source -synthetic- image into a target -real- style 
This is done -- during training -- by swapping the low-frequency component of the Fourier amplitude of the source synthetic images with that of a randomly selected target domain image.
%This is done during training, as a preprocessing step before the forward pass. 
This approach requires 1) to retrain the model for every target domain, and 2) access to an unlabelled target dataset at training time. Inspired by the use of the Fourier transform in \cite{Yang2020}, we propose a domain adaptation function $d_A$ that does not have the two requirements previously mentioned, instead, it transforms a real image $\boldsymbol{x_t}$ into a synthetic-looking image $\boldsymbol{x_s}$ following:
\begin{equation}
d_A\!\left(  \boldsymbol{x_t}, \boldsymbol{x_s} \right) = 
\left|
F^{-1} \left( t_1 \odot t_2 \odot t_3 \right)
\right|
\end{equation}
with
\begin{equation}
\begin{split}
t_1 &=  M_{h, w, \beta} |F(\boldsymbol{x_s})| \\
t_2 &= J_{h, w} - M_{h, w, \beta} |F(\boldsymbol{x_t})| \\
t_3 &= e^{
	j \tan^{-1} \left( \frac{\mathcal{I}(F(\boldsymbol{x_t}))}{\mathcal{R}(F(\boldsymbol{x_t}))} \right)
}
\end{split}
\end{equation}
where $F$ is the discrete Fourier transform, $\odot$ is the Hadamard product, $M_{h,w,\beta}$ is a matrix of dimension $h \times w$ (same as the size of the input images) with all elements equal to zero except for $M[\frac{h}{2} - \beta h:\frac{h}{2} + \beta h, \frac{w}{2} - \beta w:\frac{w}{2} + \beta w] = 1$, $J_{h,w}$ is an all-ones matrix of dimension $h \times w$, and $\mathcal{R}$ and $\mathcal{I}$ represent the real and imaginary parts respectively. In our experiments, we set the hyperparameter $\beta=0.001$, as larger values would introduce artifacts \cite{Yang2020}.
Using $d_A$, we propose to estimate the segmentation $\hat{\boldsymbol{y}}_t$ of a real image $\boldsymbol{x_t}$ following:
\begin{equation}
\label{eq:fourier_da}
\begin{split}
\hat{\boldsymbol{y}}_t &= \frac{1}{N_s} \sum_{s=1}^{N_s} f_{\boldsymbol{\theta}}
\left(
d_A \left(  \boldsymbol{x_t}, \boldsymbol{x_s} \right)
\right)
\end{split}
\end{equation}
where $N_s$ is a hyperparameter indicating the number of source domain images %semi-synthetic images 
to use for domain adaptation, $f_\theta$ is our neural network trained on images from the source domain, and $\boldsymbol{x_s}$ are randomly picked 
%semi-synthetic 
source domain images. Without loss of generality, we set $N_s=1$ in our experiments. Eq. \ref{eq:fourier_da} is only used at inference time. The results of this method are shown in tables \ref{tab:real_results} and \ref{tab:uda_results} of this supplementary material.

%\section*{A. The use of Dirichlet distribution in \textit{mix-blend}}
%\vspace{65pt}
\begin{figure*}[htb!]
	\centering
	\includegraphics[width=\textwidth]{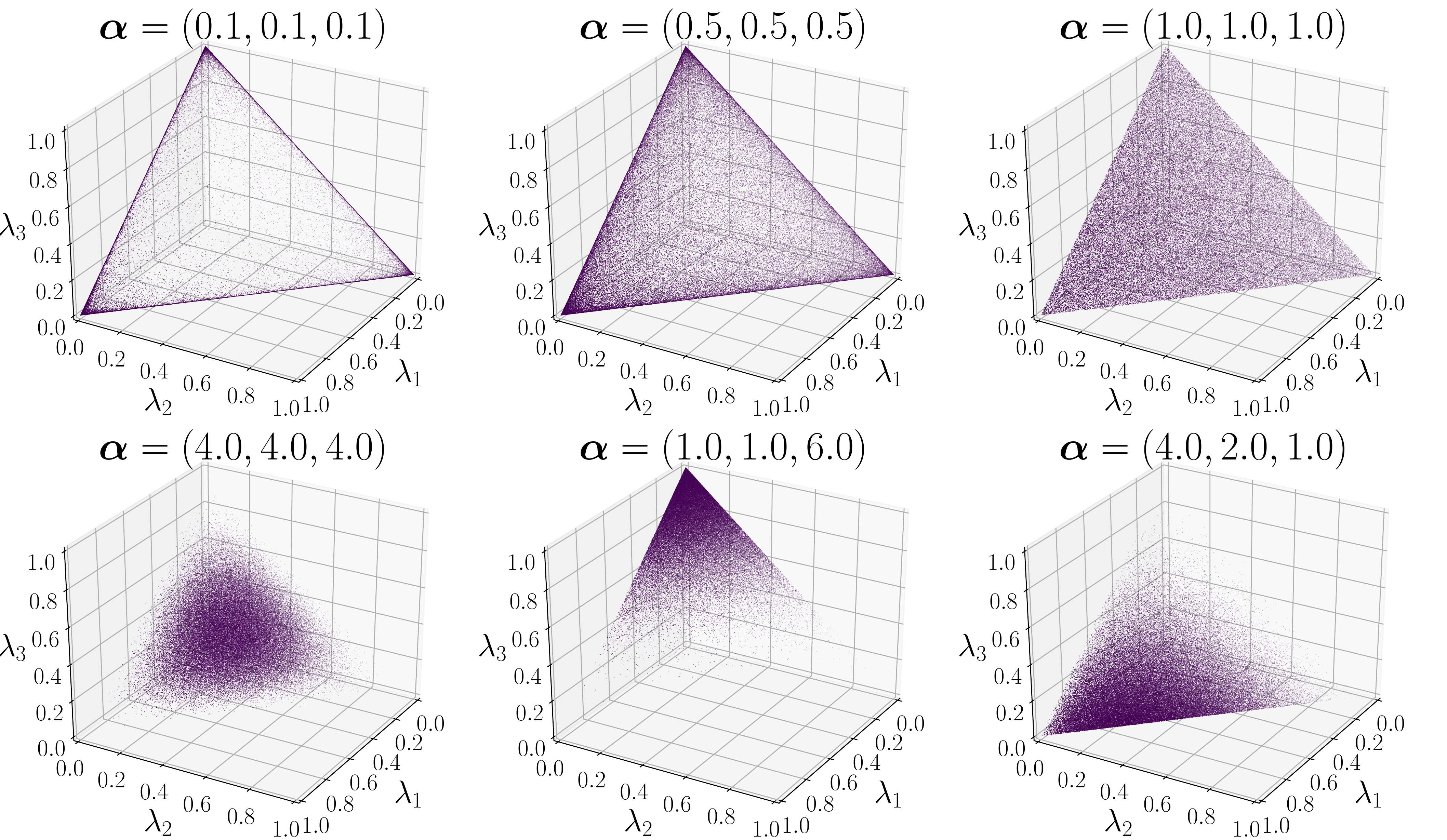}  
	\caption{A Dirichlet distribution is used to model the random weights chosen to combine the blended images in \textit{mix-blend} learning. That is, every time a pair of foreground-background images is randomly selected from the training set, M images are produced, one per blending. Different blending methods produce different plausible images. The weights employed to combine them into a single training image are randomly sampled from the distribution and later used for the weighted sum. Given $M$ blending function approximations, the Dirichlet distribution receives a parameter $\boldsymbol{\alpha} = (\alpha_1, ..., \alpha_M)$ and the samples of the distribution will be our weights $\boldsymbol{\lambda} = (\lambda_1, ..., \lambda_M)$. In our implementation we use $M=3$ blending functions, hence all the possible weight samples lie on a triangle or 2-simplex in 3D. The coordinates of every possible $\boldsymbol{\lambda}$ that belongs to the simplex sum to $1.0$, making every $\boldsymbol{\lambda}$ a vector of normalized weights that we can employ to combine the images blended with different functions. Depending on the choice of $\boldsymbol{\alpha}$, our randomly sampled weights $\boldsymbol{\lambda}$ will tend to combine the blended images in different fashions. When $\boldsymbol{\alpha} = (1.0, 1.0, 1.0)$ all linear combinations of blended images are equally likely (top right). In case that the values of $\boldsymbol{\alpha} < 1.0$ (top left), weight samples will tend to concentrate about each individual blending method or combinations of two of them. As $\boldsymbol{\alpha}$ grows, the weights samples will produce blended images that look closer to the average of $M$ blended images (bottom left).}
	\label{fig:dirichlet_samples}
\end{figure*}

\begin{figure}[htb!]
	\centering
	\includegraphics[width=\columnwidth]{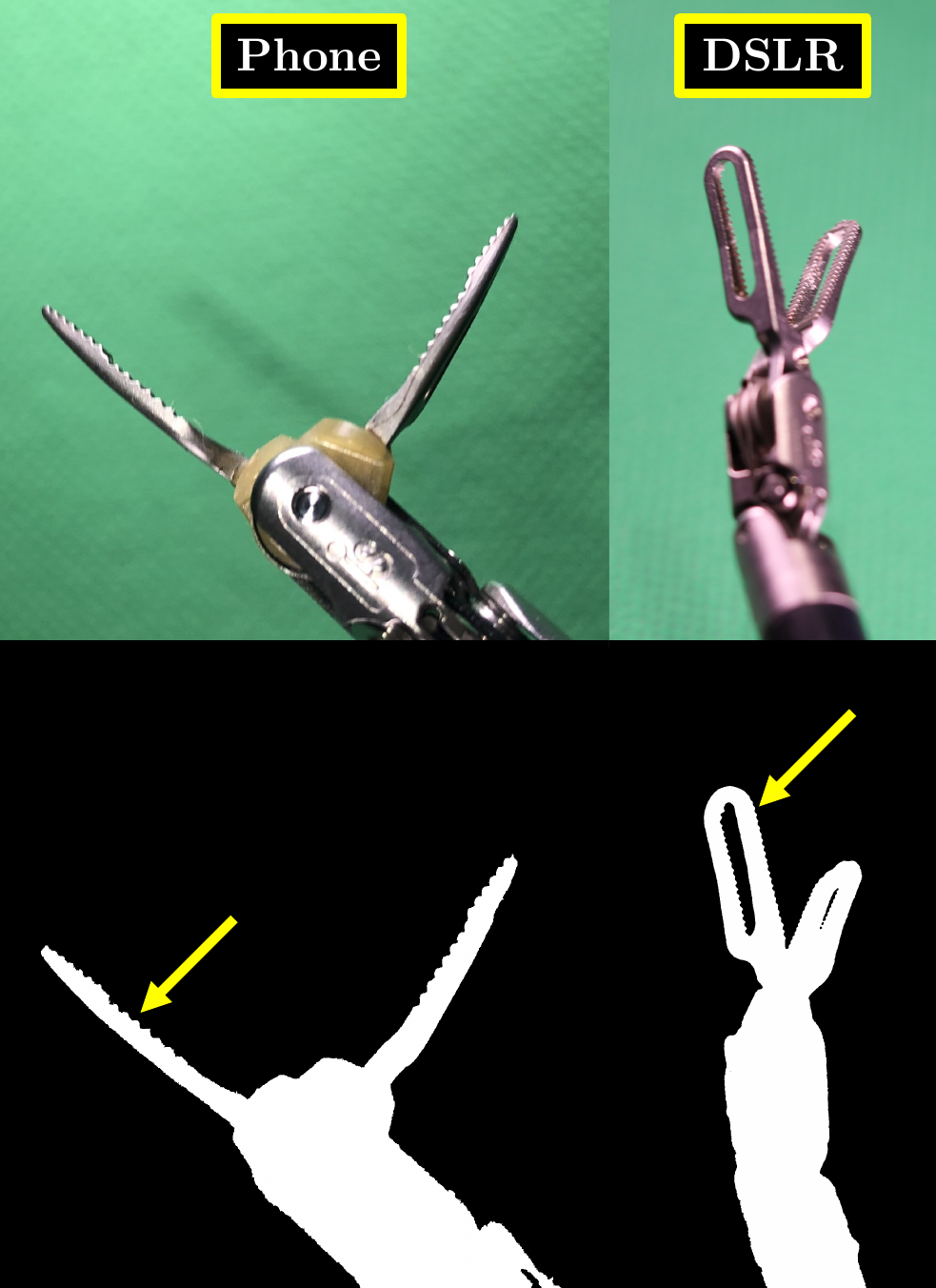}
	\caption{Frames from the foreground dataset and their automatically generated segmentation labels. 
		%Images captured by a mobile phone, and a DSLR camera. 
		%Fake blood has been spread on the instruments for some images to increase the realism of the foreground dataset.
		Figure best viewed at high resolution. High quality details such as the toothed surface (yellow arrow) captured by the segmentation in the center can then be appreciated.}
	\label{fig:fg_example}
\end{figure}

\begin{figure}[htbb!]
	\centering
	\includegraphics[width=\columnwidth]{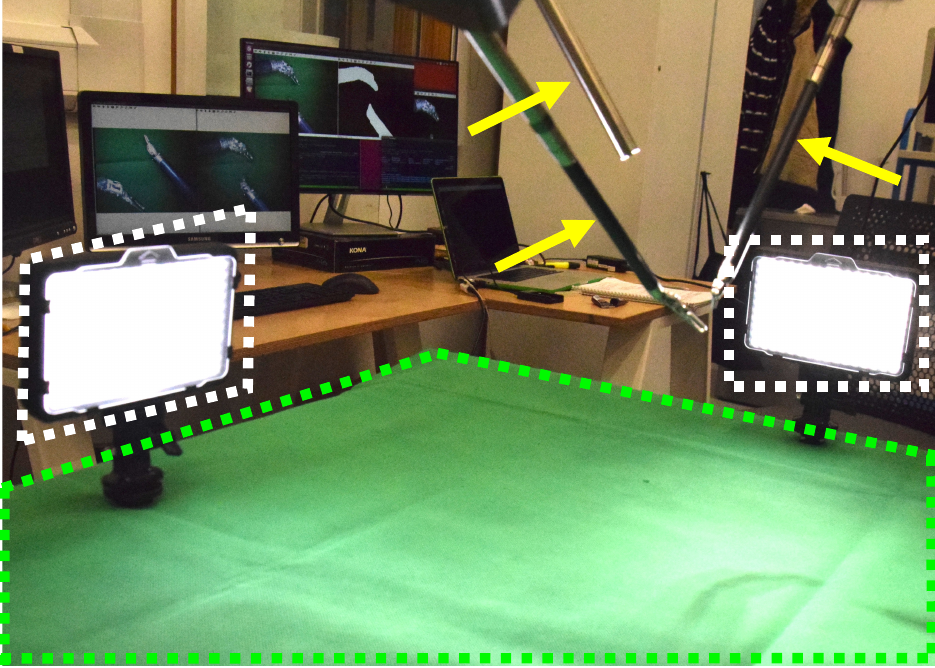}
	\caption{Recording setup. Instruments (yellow arrows), dimmable lights (white dotted line), and chroma key (green dotted line).}
	\label{fig:recording_setup}
\end{figure}

\begin{figure}[htb!]
	\centering
	\includegraphics[width=\columnwidth]{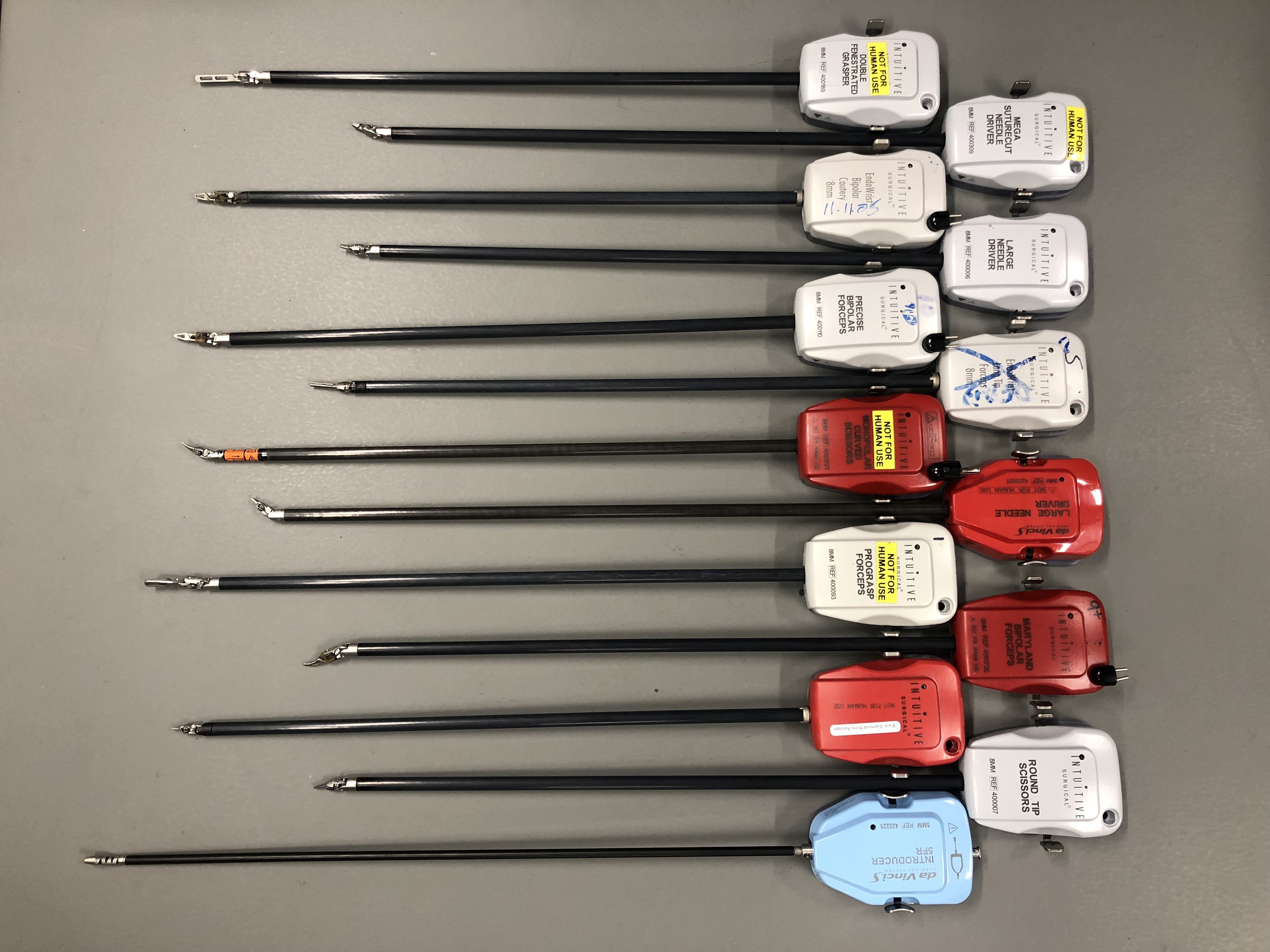}
	\caption{Instruments used for the foreground dataset.}
	\label{fig:instruments}
\end{figure}

\begin{table*}[htb!]
	\centering
	\caption{Baseline results. Training on real data and evaluating on real data. Detailed results for each EndoVis2017 video.}
	\begin{tabular}{ccccc}
		\hline
		\multicolumn{1}{c}{\bfseries Training dataset} &
		\multicolumn{1}{c}{\bfseries Domain adaptation technique} &
		\multicolumn{1}{c}{\bfseries Testing dataset} &
		\multicolumn{1}{c}{\bfseries mIoU [5\%, 95\%]} \\
		\hline
		Videos \{2, 3, 4, 5, 6, 7, 8\}  &  None  &   Video 1  &  $83.6~[71.2, 91.9]$  \\ % Exp: 144
		Videos \{1, 3, 4, 5, 6, 7, 8\}  &  None  &   Video 2  &  $89.3~[83.9, 93.3]$  \\ % Exp: 145
		Videos \{1, 2, 4, 5, 6, 7, 8\}  &  None  &   Video 3  &  $85.0~[76.7, 91.4]$  \\ % Exp: 146
		Videos \{1, 2, 3, 5, 6, 7, 8\}  &  None  &   Video 4  &  $89.8~[82.3, 94.2]$  \\ % Exp: 147
		Videos \{1, 2, 3, 4, 6, 7, 8\}  &  None  &   Video 5  &  $77.5~[63.5, 87.6]$  \\ % Exp: 148
		Videos \{1, 2, 3, 4, 5, 7, 8\}  &  None  &   Video 6  &  $85.7~[79.4, 90.4]$  \\ % Exp: 149
		Videos \{1, 2, 3, 4, 5, 6, 8\}  &  None  &   Video 7  &  $78.8~[48.0, 91.5]$  \\ % Exp: 150
		Videos \{1, 2, 3, 4, 5, 6, 7\}  &  None  &   Video 8  &  $63.4~[52.6, 76.3]$  \\ % Exp: 151
		\cline{3-4}
		&        &  Average across videos  &  $81.6~[69.7, 89.6]$ \\
	\end{tabular}
	\vspace{0.2cm}
	\label{tab:real_data_results_each_video}
\end{table*}

\begin{table*}[htb!]
	\centering
	\caption{Baseline results. Training and evaluating on manually labeled real data. No semi-synthetic data nor blending technique has been used to generate the results on this table. In the first line of the results, when training and testing on EndoVis2017, leave one out is used, following the protocol of the EndoVis2017 challenge (see section IV-E of the manuscript).}
	\begin{tabular}{lccc}
		\hline
		\multicolumn{1}{c}{\bfseries Training dataset} &
		\multicolumn{1}{c}{\bfseries Domain adaptation method} &
		\multicolumn{1}{c}{\bfseries Testing dataset} &
		\multicolumn{1}{c}{\bfseries mIoU [5\%, 95\%]} \\
		\hline
		RoboTool     &  Fourier (eq. \ref{eq:fourier_da})               &  EndoVis2017  &  $74.5~[54.7, 85.6]$  \\ % Exp: 157
		RoboTool     &  Fourier (eq. \ref{eq:fourier_da})    + GrabCut  &               &  $79.9~[51.1, 94.4]$  \\ % Exp: 157
		\hline
		EndoVis2017  &  Fourier (eq. \ref{eq:fourier_da})               &  RoboTool     &  $66.1~[40.9, 87.5]$  \\ % Exp: 152
		EndoVis2017  &  Fourier (eq. \ref{eq:fourier_da}) + GrabCut     &               &  $67.4~[34.1, 91.4]$  \\ % Exp: 152
	\end{tabular}
	\vspace{0.2cm}
	\label{tab:real_results}
\end{table*}

\begin{table*}[htb!]
	\centering
	\caption{Results of our proposed blending method in combination with different unsupervised domain adaptation techniques.}
	\begin{tabular}{lccc}
		\hline
		\multicolumn{1}{c}{\bfseries Training dataset} &
		\multicolumn{1}{c}{\bfseries Domain adaptation method} &
		\multicolumn{1}{c}{\bfseries Testing dataset} &
		\multicolumn{1}{c}{\bfseries mIoU [5\%, 95\%]} \\
		\hline
		Semi-synthetic (Mix-blend)  & Fourier (eq. \ref{eq:fourier_da})            &               &  $74.0~[58.7, 87.5]$           \\ % Exp: 138
		Semi-synthetic (Mix-blend)  & Fourier (eq. \ref{eq:fourier_da}) + GrabCut  &               &  $82.4~[61.3, 93.8]$           \\ % Exp: 138
		\hline
		Semi-synthetic (Mix-blend)  & Fourier (eq. \ref{eq:fourier_da})            &               &  $60.2~[42.4, 80.5]$           \\ % Exp: 138
		Semi-synthetic (Mix-blend)  & Fourier (eq. \ref{eq:fourier_da}) + GrabCut  &               &  $\textbf{68.9}~[43.7, 91.4]$  \\ % Exp: 138
	\end{tabular}
	\vspace{0.2cm}
	\label{tab:uda_results}
\end{table*}

\begin{figure*}[htb!]
	\includegraphics[width=\textwidth]{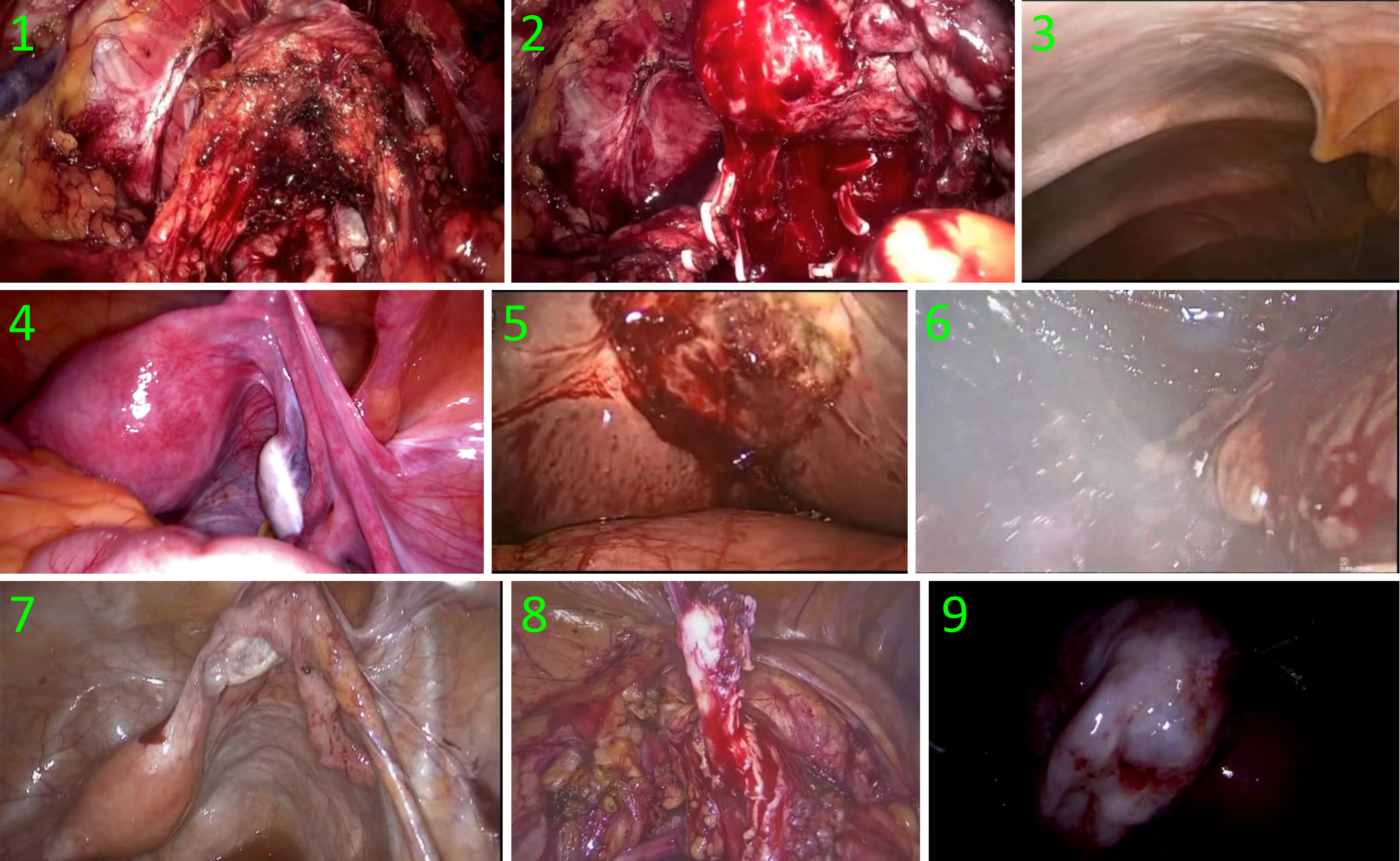}
	\caption{Images from the background dataset displaying changes due to the presence of tools. We can observe tool-tissue interactions, blood, debris, smoke and lighting artifacts due to tissue manipulation. In sub-image $1$, the tissue has been manipulated by thermocautery. In sub-images $2$ and $5$, blood is shown as a result of surgical cuts. Sub-image $3$ displays a surgical trocar about to be inserted. Sub-images $4$, $7$, and $8$ show tissue being pulled. In sub-image $6$, surgical smoke is disrupting the view. Sub-image $9$ has large areas of shadows created by the chunk of tissue being manipulated.}
	\label{fig:background_tool_tissue_interactions}
\end{figure*}

\begin{figure*}[htb!]
	\centering
	\includegraphics[width=\textwidth]{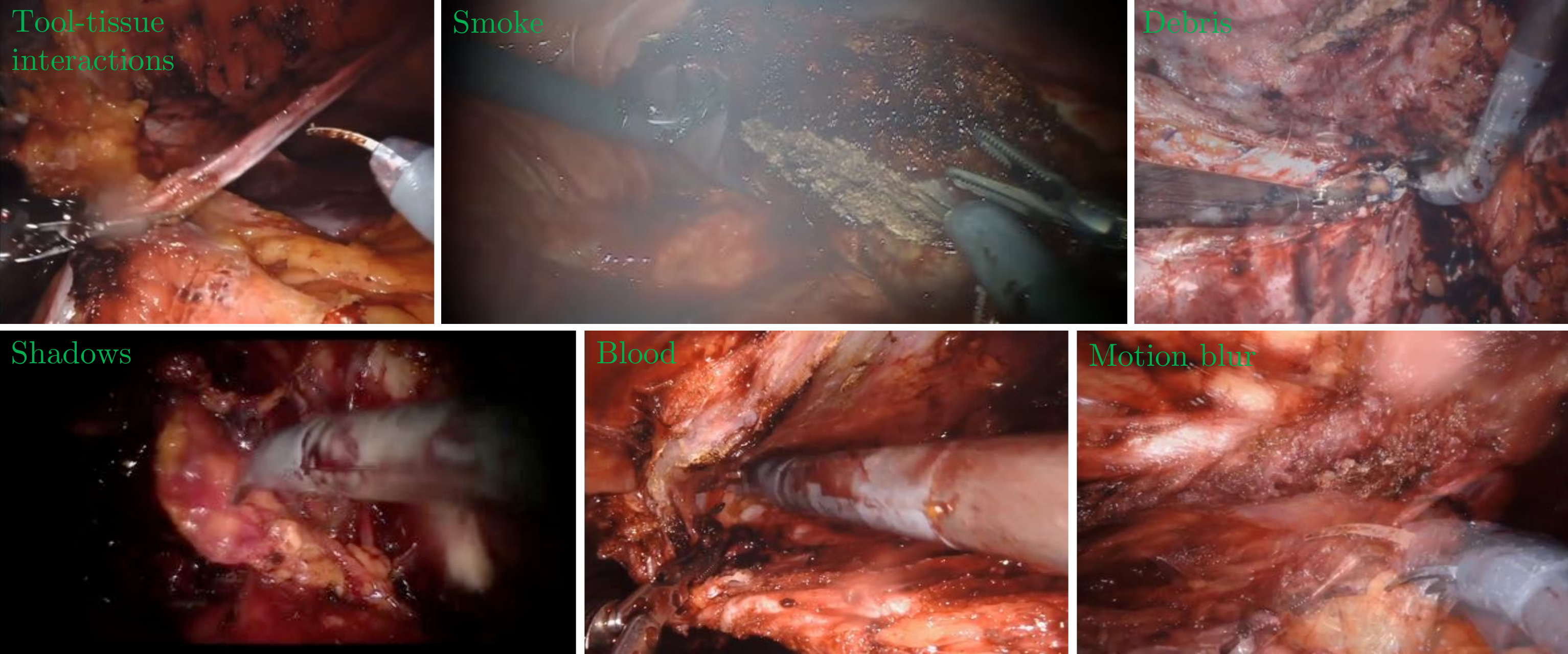}
	\caption{
		Exemplar real clinical images of the RoboTool testing set displaying environments modified by the presence of instruments (tool-tissue interaction) and indirectly (smoke, blood, debris, shadows).
	}
	\label{fig:robotool}
\end{figure*}

\begin{figure*}[htb!]
	\centering
	\includegraphics[width=\textwidth]{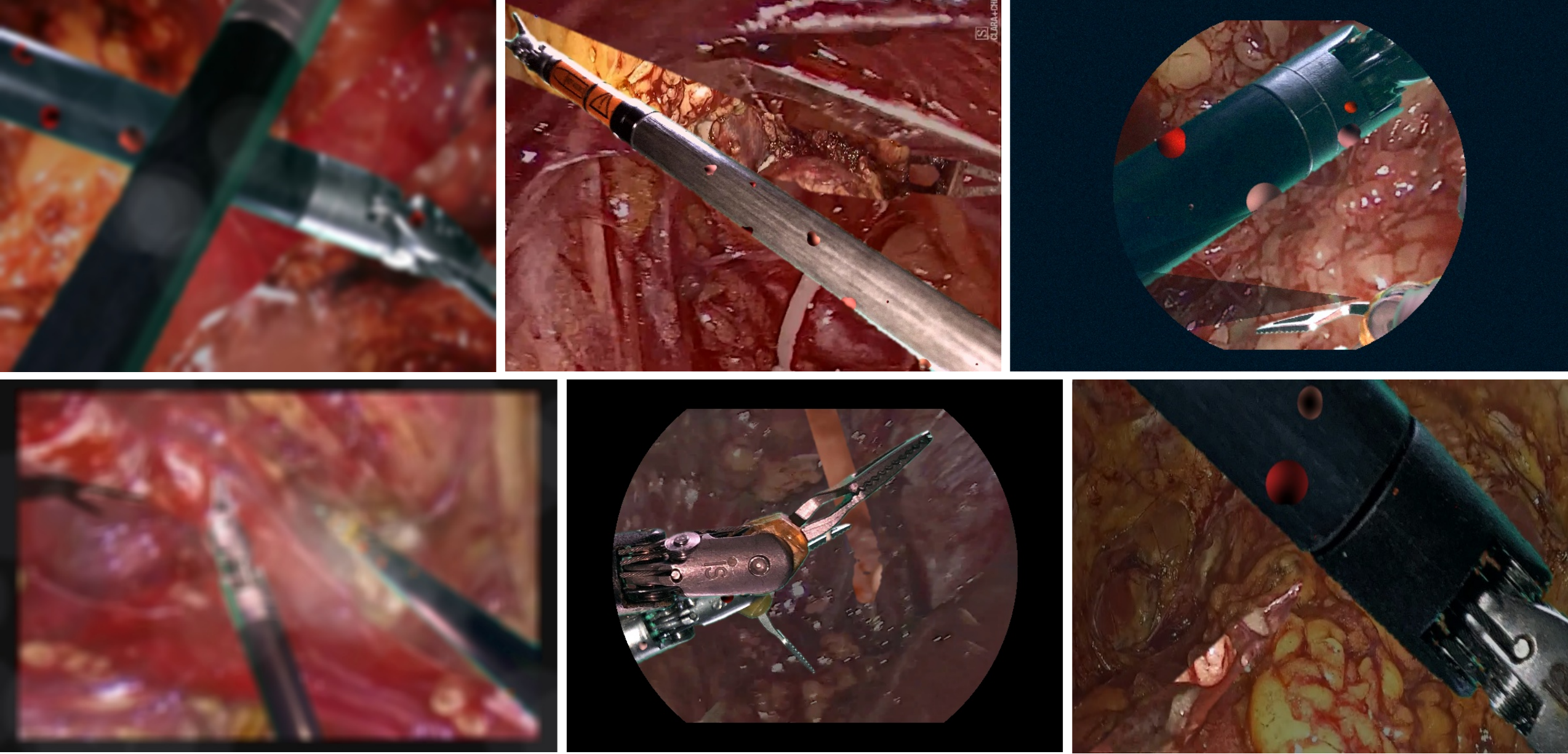}
	\caption{
		Exemplar semi-synthetic training images containing a variety of synthetic artifacts. Blood droplets and small pieces of tissue debris (present in all the images shown). Synthetic smoke (top-left and bottom-left). Different border frames (top-right, bottom-left, bottom-center). Synthetic shadows (top-right).
	}
	\label{fig:blended}
\end{figure*}

\begin{figure*}[htb!]
	\includegraphics[width=\textwidth]{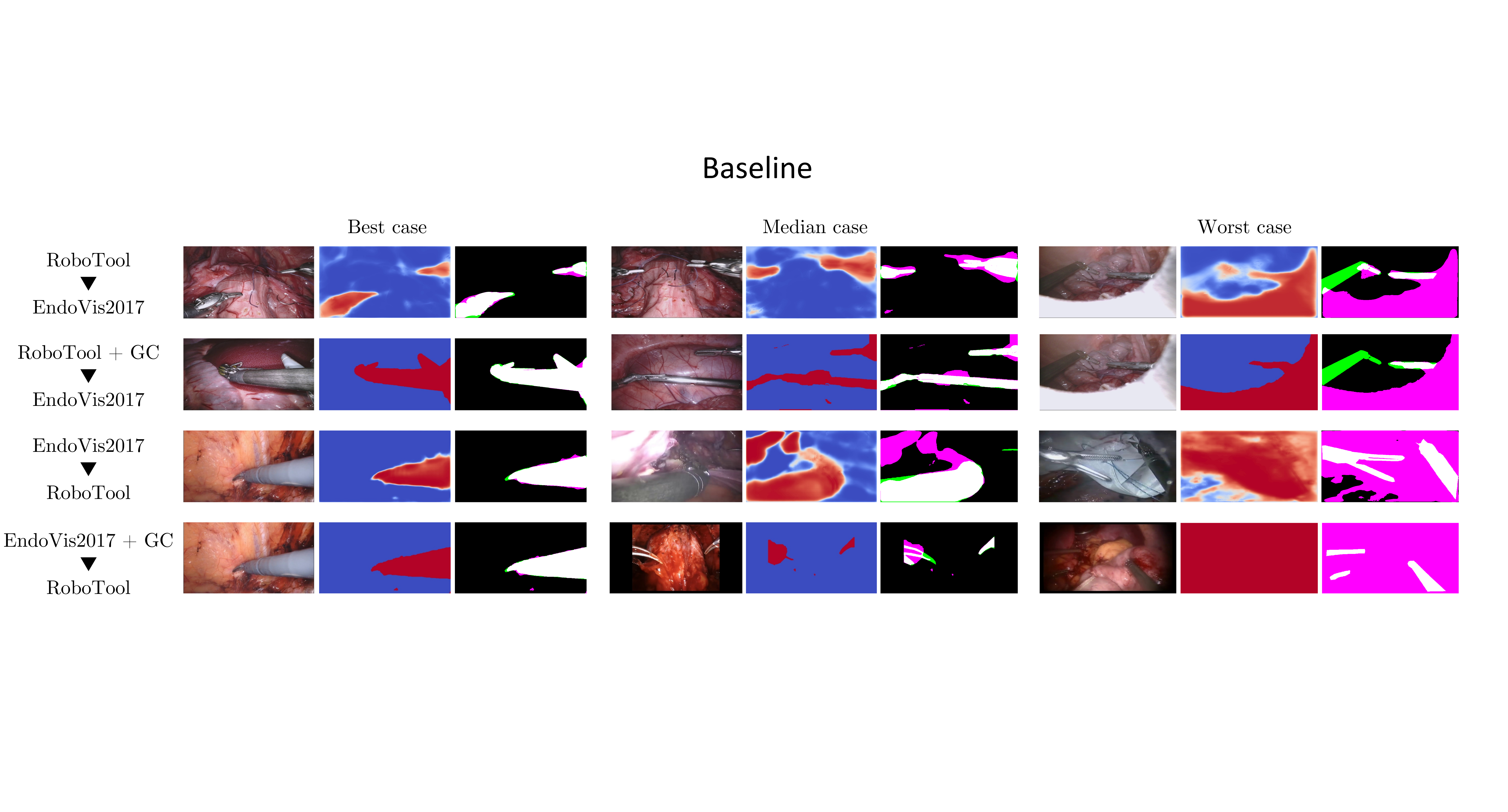}
	\caption{Baseline results (not including any semi-synthetic data). Training and evaluating on manually labeled real data. The name of the dataset used for training is placed on top and the name of the testing dataset is specified under the arrow. GC stands for GrabCut. When specified, the results have been postprocessed by GrabCut.}
	\label{fig:best_median_worst_baseline}
\end{figure*}
\begin{figure*}[htb!]
	\includegraphics[width=\textwidth]{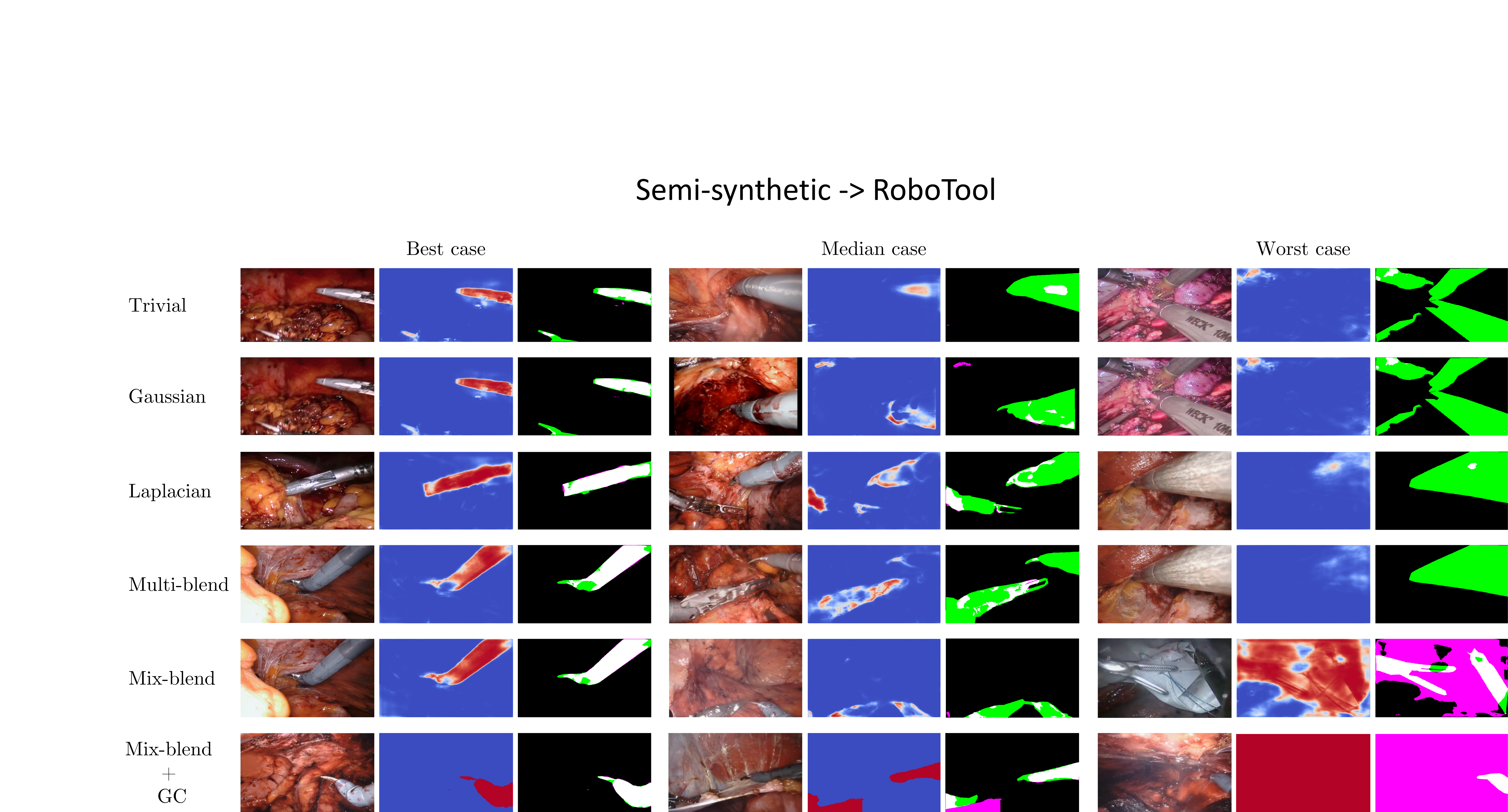}
	\caption{Results of the different methods when training on semi-synthetic data and testing on RoboTool real clinical data.}
	\label{fig:best_median_worst_robotool}
\end{figure*}
\begin{figure*}[htb!]
	\includegraphics[width=\textwidth]{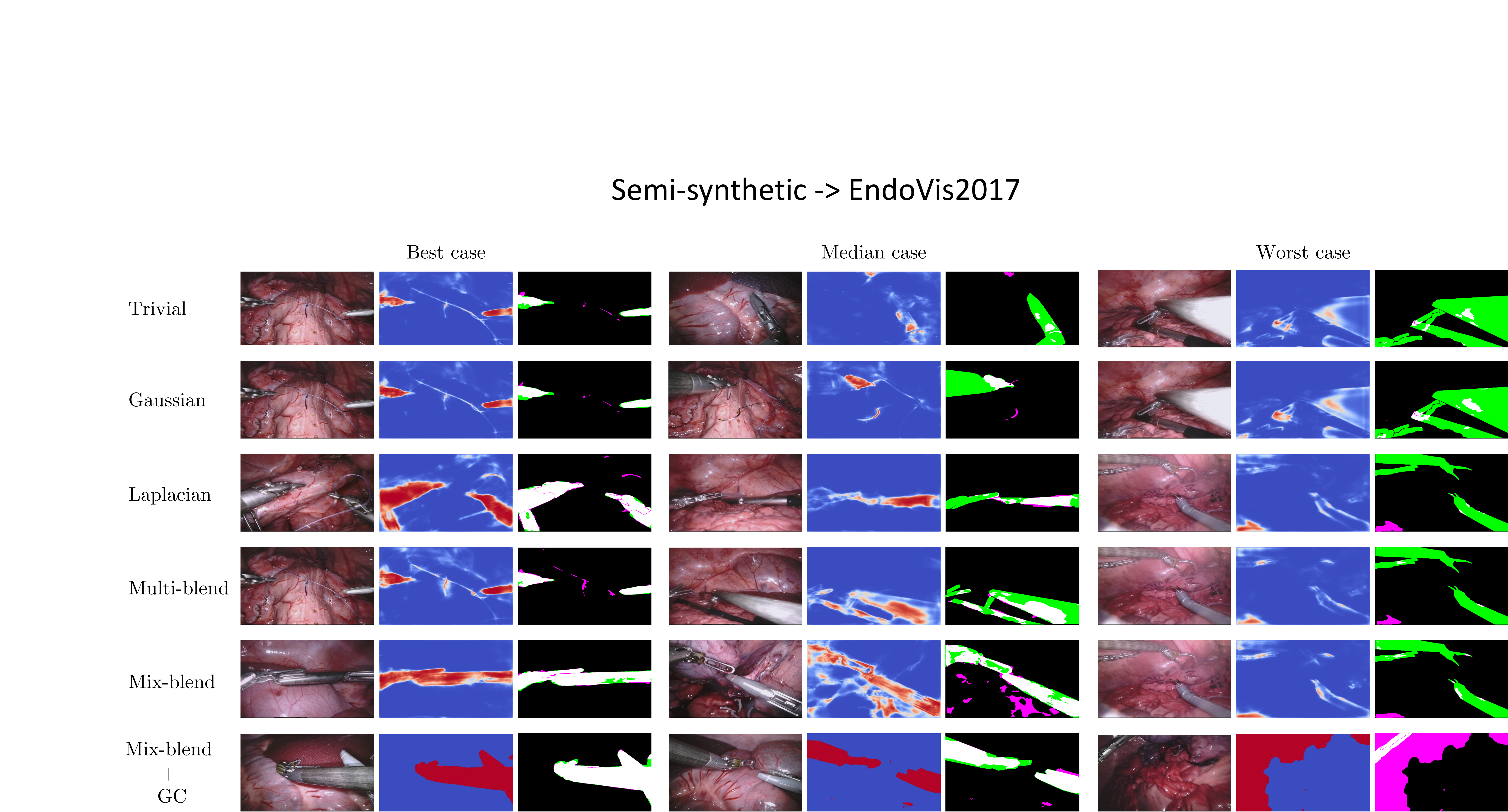}
	\caption{Results of the different methods when training on semi-synthetic data and testing on EndoVis2017 real \textit{ex vivo} data.}
	\label{fig:best_median_worst_endovis}
\end{figure*}

\end{document}